\documentclass[Afour, sagev, times]{sagej}

\usepackage{moreverb,url}
\usepackage{subcaption}
\usepackage{hyperref}
\usepackage{multirow}
\usepackage{multicol}
\usepackage{mathtools}
\usepackage{amsmath}
\newcommand\BibTeX{{\rmfamily B\kern-.05em \textsc{i\kern-.025em b}\kern-.08em
T\kern-.1667em\lower.7ex\hbox{E}\kern-.125emX}}

\usepackage{dblfloatfix}
\usepackage{amssymb}
\def\volumeyear{2022}

\begin{document}

\runninghead{Hlaing et al.}

\title{Farm-wide virtual load monitoring for offshore wind structures via Bayesian neural networks}

\author{Nandar Hlaing\affilnum{1}, Pablo G. Morato\affilnum{2}, Francisco de Nolasco Santos\affilnum{3}, Wout Weijtjens\affilnum{3}, \\ Christof Devriendt\affilnum{3} and Philippe Rigo\affilnum{1}}

\affiliation{\affilnum{1}ANAST, University of Liege, 4000 Liege, Belgium\\
\affilnum{2}Department of Wind and Energy Systems, Technical University of Denmark, 4000 Roskilde, Denmark\\
\affilnum{3}OWI-Lab, Vrije Universiteit Brussel, 1050 Brussels, Belgium}

\corrauth{Nandar Hlaing, ANAST, University of Liege, 4000 Liege, Belgium.}

\email{nandar.hlaing@uliege.be}

\begin{abstract}

Offshore wind structures are exposed to a harsh marine environment and are subject to deterioration mechanisms throughout their operational lifetime. Even if the deterioration evolution of structural elements can be estimated through physics-based deterioration models, the uncertainties involved in the process hurdle the selection of lifecycle management decisions, e.g., lifetime extension. In this scenario, the collection of relevant information through an efficient monitoring system enables the reduction of uncertainties, ultimately driving more optimal lifecycle decisions. However, a full monitoring instrumentation implemented on all wind turbines in a farm might become unfeasible due to practical and economical constraints. Besides, certain load monitoring systems often become defective after a few years of marine environment exposure. Addressing the aforementioned concerns, a farm-wide virtual load monitoring scheme directed by a fleet-leader wind turbine offers an attractive solution. Fetched with data retrieved from a fully instrumented wind turbine, a model can be firstly trained and then deployed, thus yielding load predictions of non-fully monitored wind turbines, from which only standard data remains available, e.g., SCADA. During its deployment stage, the pretrained virtual monitoring model might, however, receive previously unseen monitoring data, thus often producing inaccurate load predictions. In this paper, we propose a virtual load monitoring framework formulated via Bayesian neural networks (BNNs) and we provide relevant implementation details needed for the construction, training, and deployment of BNN data-based virtual monitoring models. As opposed to their deterministic counterparts, BNNs intrinsically announce the uncertainties associated with generated load predictions and allow to detect inaccurate load estimations generated for non-fully monitored wind turbines. The proposed virtual load monitoring is thoroughly tested through an experimental campaign in an operational offshore wind farm and the results demonstrate the effectiveness of BNN models for `fleet-leader’-based farm-wide virtual monitoring.

\end{abstract}


\maketitle
% \section{Nomenclature}
% \newpage
\section{Introduction}
Offshore wind turbines are continuously exposed to a combined wind and wave load excitation, thus inducing fatigue deterioration and other mechanical stressors throughout their service life. The evolution of fatigue damage can be estimated through physics-based engineering models, yet the resulting deterioration predictions contain significant uncertainties. Combined with engineering models, manual and/or robotic inspections can be conducted in order to reduce the uncertainties associated with deterioration estimations, hence supporting more rational and informed maintenance decisions. \cite{Farhan2022, Hlaing2022a} With the advent of modern sensor technologies, monitoring systems are increasingly being deployed with the objective of continuously monitoring the deterioration experienced by offshore wind structures, thus also enabling decision-makers to make timely and informed decisions. \cite{Zhao2019, Nielsen2022} For example, fatigue load monitoring through strain gauges provides valuable information that can be used to estimate the remaining useful fatigue lifetime \cite{Schedat2016, Mai2019, Pacheco2022} and/or to update probabilistically modeled time-varying deterioration mechanisms. \cite{Smarsly2013, Morato2021, Hlaing2022}

However, strain gauges, and other monitoring systems, are also prone to deterioration in a harsh marine environment and their operational lifespan is normally shorter than the service life considered for an offshore wind turbine. In this context, virtual load monitoring, either physics- or data-based, offers an adequate solution, providing load information once strain sensors are no longer functional. \cite{Cosack2020} Each approach features its own advantages and disadvantages, and the choice should be mainly based on one's knowledge in physics and on data availability.\cite{Karniadakis2021} Physics-based approaches demand the development of a numerical model in order to simulate the loads under actual environmental and operational conditions (EOCs), which are collected via a supervisory control and data acquisition (SCADA) system. Based on collected on-site structural response information, physics-based models can be periodically updated. \cite{Xu2020, Zhang2021} Whereas physics-based approaches require less data than fully data-based schemes, the development and implementation of complex dynamical models demand specific knowledge, e.g., thrust coefficient, soil parameters, that is usually hard to retrieve in real case scenarios.

Alternatively, one can benefit from easily-accessible data, e.g., SCADA data, accelerometers, among others, to train a deterministic or probabilistic data-based model.\cite{Maes2016, Hlaing2022} The majority of previous research studies investigate the application of deterministic models for virtually monitoring offshore wind turbines, mostly focusing on learning relationships between SCADA data and fatigue loads. \cite{Perez2018, Santos2020} In a few cases, high frequency acceleration measurements are combined with SCADA data in order to adequately capture dynamic load components. \cite{Noppe2016} Generally, deterministic models do not explicitly indicate the uncertainty associated with the generated predictions, except for model verification analysis with respect to sensor data.\cite{Kennedy2001, Hlaing2022} On the other hand, probabilistic approaches can intrinsically provide an indication of the uncertainty contained in the produced predictions. From the limited number of reported probabilistic virtual monitoring methods, \cite{Pimenta2022,Zou2023} Singh et al. \cite{Singh2022} investigated the applicability of heteroscedastic Gaussian processes for probabilistically modelling turbine loads, and similarly, a Gaussian process-based method was also proposed with the objective of extrapolating monitored fatigue damage from an instrumented tower section to any other level, based on a covariance matrix defined via acceleration signals. \cite{Pimenta2022}  

An additional challenge faced by virtual load monitoring schemes is their applicability at a farm-wide level. In this perspective, physics-based approaches can easily become burdensome since a complex numerical model of the entire wind farm is needed, where each turbine structural design is tailored to a specific water depth, soil conditions, etc. More recently, certain data-based virtual monitoring schemes have adopted a fleet-leader approach, in which load measurements collected from monitored turbines are extrapolated to non-instrumented ones. For instance, a recent study extrapolates short-term damage measurements from an instrumented fleet-leader to the wind farm relying on binned SCADA data and turbine conditions (i.e., operational or parked). \cite{Noppe2020} However, other relevant information for the estimation of fatigue damage evolution, e.g., structural dynamics variations among wind turbines, might not be appropriately captured in SCADA data. \cite{Weijtens2016} In this regard, data-based models that combine both SCADA data and response information collected from accelerometers can provide farm-wide fatigue predictions more effectively. \cite{Santos2022}

As previously explained during the literature survey, recent investigations often rely on deterministic data-based approaches for the formulation of virtual load monitoring models, yet their inability to detect potential conflicts during the deployment stage, e.g., inaccurate load predictions when the network is fed with previously unseen input data, limits their applicability to farm-wide virtual monitoring implementations. Unless additional sensors are installed with the objective of retrieving `ground truth’ load measurements, the model uncertainty associated with predictions generated for non-fully monitored wind turbines cannot be quantified. In this paper, we cast a virtual load monitoring framework that relies on Bayesian neural networks and probabilistic deep learning methods in order to provide farm-wide load predictions and enable the intrinsic quantification of aleatory uncertainty (emerging due to the random nature of the physical system) and epistemic uncertainty (arising due to lack of knowledge of the system). The proposed virtual monitoring framework is tested on a dataset collected from three wind turbines that are currently operating in a Belgian offshore wind farm. In particular, a Bayesian neural network is trained based on SCADA and accelerometer data (inputs) along with the corresponding damage equivalent moments (labels) collected from a specific offshore wind turbine. The reduction of model uncertainty with increasing dataset size is thoroughly quantified and the resulting Bayesian model is cross-validated for the same turbine as well as for other turbines located in the same wind farm. Interestingly, the results showcase that Bayesian models are able to intrinsically report higher model uncertainties when tested on a wind turbine characterized with a dynamic behavior different from the one employed during the training, thus demonstrating the ability of the proposed virtual load monitoring scheme to automatically inform if the provided predictions are inaccurate and whether further information collection actions are needed.

\section{Bayesian neural networks}

Most recent applications of Bayesian neural networks for offshore wind energy settings have mainly focused on wind speed and power forecasting. \cite{Mbuvha2017, Wang2017, Mbuvha2021} Up to the knowledge of the authors, Bayesian neural networks-based structural health monitoring methods have not yet been formally proposed for offshore wind applications. In this section, we briefly introduce Bayesian neural networks from a general theoretical perspective, since this will facilitate the application and understanding of the proposed virtual monitoring framework. In essence, a Bayesian neural network (BNN) is a stochastic artificial neural network trained via Bayesian inference, featuring the combined strength of deep learning and Bayesian theory in order to provide a rich probabilistic interpretation of the generated predictions.

\begin{figure*}
     \centering
     \begin{subfigure}[b]{0.49\textwidth}
         \centering
         \includegraphics{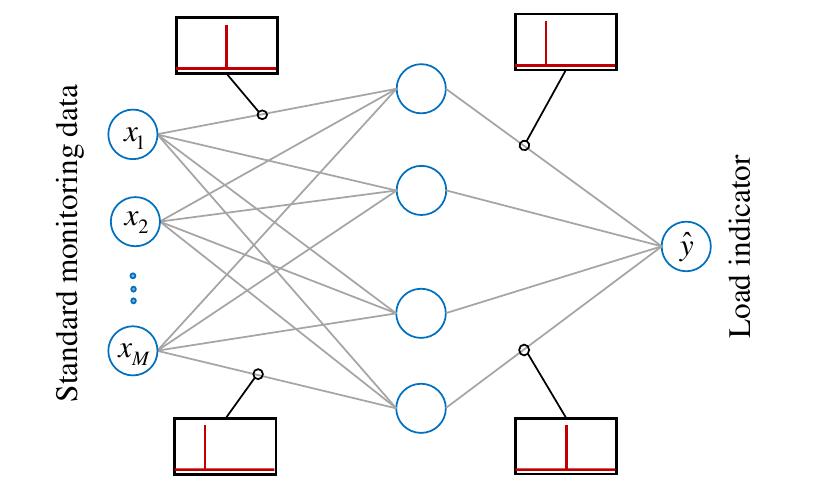}
         \caption{Deterministic neural network with weights and biases specified as point estimates $\boldsymbol{\theta}$ along with a deterministic output.}
     \end{subfigure}
     \hfill
     \begin{subfigure}[b]{0.49\textwidth}
         \centering
         \includegraphics{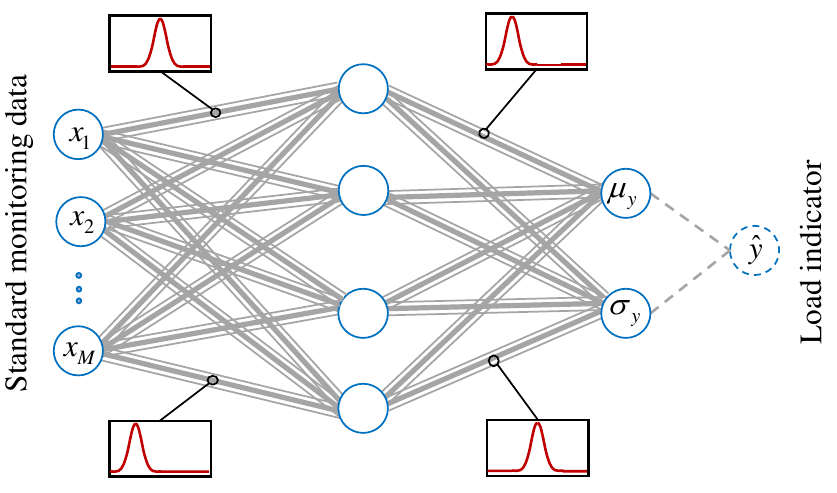}
         \caption{Bayesian neural network with weights and biases specified as probability distributions $P(\boldsymbol{\theta})$ along with a stochastic output.}
     \label{fig:BNN}
     \end{subfigure}
     \caption{Schematic diagrams comparing the topology and constituents of a standard deterministic neural network (DNN) and a Bayesian neural network (BNN), both mapping standard input monitoring data $\boldsymbol{x}$ to a load indicator $y$.}
     \label{fig:DNN_BNN}
\end{figure*}

The key defining characteristic of Bayesian neural networks with respect to conventional artificial neural networks (ANNs) is their stochastic neural network components, i.e., stochastic activations and/or weights, \cite{Jospin2022} as illustrated in Figure \ref{fig:DNN_BNN}, enabling this way multiple model parametrizations $\boldsymbol{\theta}$, each of them associated with a probability distribution $P(\boldsymbol{\theta})$. In most applications, the output prediction of a BNN is also formulated as a probability distribution, thereby quantifying uncertainties associated with the underlying process, i.e., aleatory uncertainties that naturally arise from random physical phenomena or inherent noise in the training data, and model (epistemic) uncertainty due to the limited information used for training the networks. Whereas aleatory uncertainty is irreducible, model uncertainty can be reduced as the networks learn from additionally considered training data. In contrast to their deterministic counterparts, BNNs report high epistemic uncertainty in regions where only a few (or none) training points are available.

When modeling BNNs, and similarly to ANNs, the selection of the network architecture plays a key role. Various ANNs’ topologies commonly used in machine learning applications, e.g., feed-forward, convolutional, and recurrent neural networks, are also applicable to BNNs. Additionally, a stochastic model should also be defined, i.e., a prior distribution of model parameters $P(\boldsymbol{\theta})$. While the choice of prior distributions is arbitrary, a Gaussian prior with zero mean and diagonal covariance $\mathcal{N}(\boldsymbol{0},\sigma \boldsymbol{I})$ is commonly adopted. Gaussian priors are often preferred due to their advantageous mathematical properties, e.g., its logarithmic formulation is the cornerstone of most learning algorithms.

\subsection{Inference Methods for BNNs} 

Conditioned to the training dataset, $D$, the posterior probability distribution of neural network weights $P(\boldsymbol{\theta} \mid D)$ can be computed via Bayes' theorem:
\begin{equation}
    P(\boldsymbol{\boldsymbol{\theta}} \mid D) = \frac{P(D \mid \boldsymbol{\boldsymbol{\theta}}) P(\boldsymbol{\theta})}{\int {P(D \mid \boldsymbol{\theta}) P(\boldsymbol{\theta}) d\boldsymbol{\theta}}}.
    \label{eq:Bayes}
\end{equation}
The calculation of the posterior distribution is usually intractable for continuous probabilistic settings. Therefore, various methods have been developed in order to estimate the Bayesian posterior, e.g., Laplace approximation, \cite{Ritter2018} variational inference, \cite{Osawa2019} Markov Chain Monte Carlo sampling. \cite{wenzel2020}

Among various Markov chain Monte Carlo (MCMC) algorithms, Metropolis-Hastings \cite{Metropolis1953, Hastings1970} has been widely used in Bayesian statistics,\cite{Robert1999} benefiting from the fact that only a proportional distribution to the posterior is needed.
Despite MCMC algorithms enable the estimation of posterior distributions through sampling processes, their applicability is still limited to small datasets and medium complex models, e.g., 10 to 100 variables. Alternatively, variational inference (VI) has been widely used for settings featuring large datasets and highly complex models with thousands to millions of parameters and can be applied to most neural network architectures.
\cite{Graves2011} The interested reader is directed to Jospin et al. \cite{Jospin2022} for a detailed overview of VI and other inference methods applicable to Bayesian neural networks. 

The objective of variational inference (VI) is to approximate the potentially complex posterior distribution of weights by a simpler one, denoted as variational distribution. Gaussian distributions are often used to estimate the posteriors, whose parameters $\lambda = (\mu_{\boldsymbol{\theta}}, \sigma_{\boldsymbol{\theta}})$ are, therefore, commonly known as variational parameters. VI methods adjust $\lambda$ so that the variational distribution $q_{\lambda}(\boldsymbol{\theta})$ closely resembles the posterior $P(\boldsymbol{\theta} \mid D)$. 
The similarity or divergence between the two distributions is formally described by the Kullback-Leibler (KL) divergence,\cite{Kullback1951} a non-symmetric and information-theoretic measure of the statistical difference between two probability distributions. An optimal solution for the variational distribution $q_{\lambda}(\boldsymbol{\theta})$ is then obtained by minimizing KL divergence between $q_{\lambda}(\boldsymbol{\theta})$ and the posterior $P(\boldsymbol{\theta}|D)$. Mathematically, KL corresponds to the expected value of the difference between the logarithmic probabilities associated with the two distributions:

\begin{align}
    \MoveEqLeft[2] \mathbb{KL}\left(q_{\lambda} (\boldsymbol{\theta}) \mid \mid P(\boldsymbol{\theta}\mid D)\right) & \\
    &= \int{q_{\lambda} (\boldsymbol{\theta})\log\frac{q_{\lambda} (\boldsymbol{\theta})}{P(\boldsymbol{\theta}\mid D)}d\boldsymbol{\theta}} \notag\\
    &= \int{q_{\lambda} (\boldsymbol{\theta})\log\frac{q_{\lambda} (\boldsymbol{\theta})}{\frac{P(\boldsymbol{\theta} ,D)}{P(D)}}d\boldsymbol{\theta}} \notag \\
    &= \int{q_{\lambda} (\boldsymbol{\theta}) \log P(D)d\boldsymbol{\theta}} - \int{q_{\lambda} (\boldsymbol{\theta})\log\frac{P(\boldsymbol{\theta} ,D)}{q_{\lambda} (\boldsymbol{\theta})}d\boldsymbol{\theta}} \notag \\
    &=  \log P(D) - \int{q_{\lambda} (\boldsymbol{\theta})\log\frac{P(D \mid \boldsymbol{\theta})P(\boldsymbol{\theta})}{q_{\lambda} (\boldsymbol{\theta})}d\boldsymbol{\theta}}. \notag 
\end{align}
    % & \quad +o+p+q+r+s+t+u+v+w+x+y+z+1+2+3+4 \notag\\
% \begin{multline}
%  \mathbb{KL}\left(q_{\lambda} (\boldsymbol{\theta}) \mid \mid P(\boldsymbol{\theta}\mid D)\right)
% % \begin{aligned}
%     &= \int{q_{\lambda} (\boldsymbol{\theta})\log\frac{q_{\lambda} (\boldsymbol{\theta})}{P(\boldsymbol{\theta}\mid D)}d\boldsymbol{\theta}} \\
%     \\
%     &= \int{q_{\lambda} (\boldsymbol{\theta})\log\frac{q_{\lambda} (\boldsymbol{\theta})}{\frac{P(\boldsymbol{\theta} ,D)}{P(D)}}d\boldsymbol{\theta}}  \\
%     \\
%     &= \int{q_{\lambda} (\boldsymbol{\theta}) \log P(D)d\boldsymbol{\theta}} - \int{q_{\lambda} (\boldsymbol{\theta})\log\frac{P(\boldsymbol{\theta} ,D)}{q_{\lambda} (\boldsymbol{\theta})}d\boldsymbol{\theta}}\\
%     \\
%     &=  \log P(D) - \int{q_{\lambda} (\boldsymbol{\theta})\log\frac{P(D \mid \boldsymbol{\theta})P(\boldsymbol{\theta})}{q_{\lambda} (\boldsymbol{\theta})}d\boldsymbol{\theta}}\\
% % \end{aligned}
% \end{multline}
Since the first term is independent with respect to the variational parameters, minimizing $\mathbb{KL}\left(q_{\lambda} (\boldsymbol{\theta}) \mid \mid P(\boldsymbol{\theta}\mid D)\right)$ is equivalent to maximizing the second term, often denoted as the evidence lower bound objective (ELBO): %or minimizing its negation:
\begin{flalign}
    \text{ELBO} &= \int{q_{\lambda} (\boldsymbol{\theta})\log\frac{P(D \mid \boldsymbol{\theta})P(\boldsymbol{\theta})}{q_{\lambda} (\boldsymbol{\theta})}d\boldsymbol{\theta}} & \\
    &= -\int{q_{\lambda} (\boldsymbol{\theta})\log\frac{q_{\lambda} (\boldsymbol{\theta})}{P(\boldsymbol{\theta})}d\boldsymbol{\theta}}  \notag\\ 
    &+ \int{q_{\lambda}(\boldsymbol{\theta})\log P(D \mid \boldsymbol{\theta} )d\boldsymbol{\theta}}. \notag 
\end{flalign}
% \notag \\& \quad
In particular, the loss function to be minimized corresponds to the negative ELBO and the optimal variational parameter $\lambda^*$ can be, therefore, formulated as:
% \begin{multline}
%   \lambda^* = \text{argmin}\{\mathbb{KL}\left(q_{\lambda} (\boldsymbol{\theta}) \mid \mid P(\boldsymbol{\theta})\right) \\
%   - \mathbb{E}_{\boldsymbol{\theta} \sim q_{\lambda}}\left[\log {(P(D\mid \boldsymbol{\theta}))}\right]\}.
%     \label{eq:Loss}
% \end{multline}
\begin{multline}
  \lambda^* = \text{argmin}\{\mathbb{KL}\left(q_{\lambda} (\boldsymbol{\theta}) \mid \mid P(\boldsymbol{\theta})\right) \\
  - \mathbb{E}_{\boldsymbol{\theta} \sim q_{\lambda}}\left[\log {(P(D\mid \boldsymbol{\theta}))}\right]\}.
    \label{eq:Loss}
\end{multline}
The first term represents the KL divergence between the variational distribution $q_{\lambda}(\boldsymbol{\theta})$ and the known prior $P(\boldsymbol{\theta})$ and it makes sure the variational distribution is close to the prior distribution.  When priors are selected with zero mean, minimizing $\mathbb{KL}\left(q_{\lambda} (\boldsymbol{\theta})\mid \mid P(\boldsymbol{\theta})\right)$ resembles the concept of regularization, i.e., driving weight estimates toward zero. \cite{James2013} 
% In case of Gaussian priors $\mathcal{N}(\mu_1,\sigma_1^2)$ and variational posteriors $\mathcal{N}(\mu_2,\sigma_2^2)$, the KL divergence can be computed as follows:
% \begin{multline}
%     \mathbb{KL}\left(\mathcal{N}(\mu_1,\sigma_1^2)\mid\mid\mathcal{N}(\mu_2,\sigma_2^2)\right) = \log \frac{\sigma_2}{\sigma_1}\\
%     +\frac{\sigma_1^2 +(\mu_1-\mu_2)^2}{2\sigma_2^2}-\frac{1}{2}.
% \end{multline}
The second term of Equation (\ref{eq:Loss}) 
computes the expected negative log-likelihood of the training data given the weight $\boldsymbol{\theta}$ distributed according to $q_{\lambda} (\boldsymbol{\theta})$. Minimizing this term controls that BNN’s produced predictions match training target data.

\begin{figure}
    \centering   \includegraphics{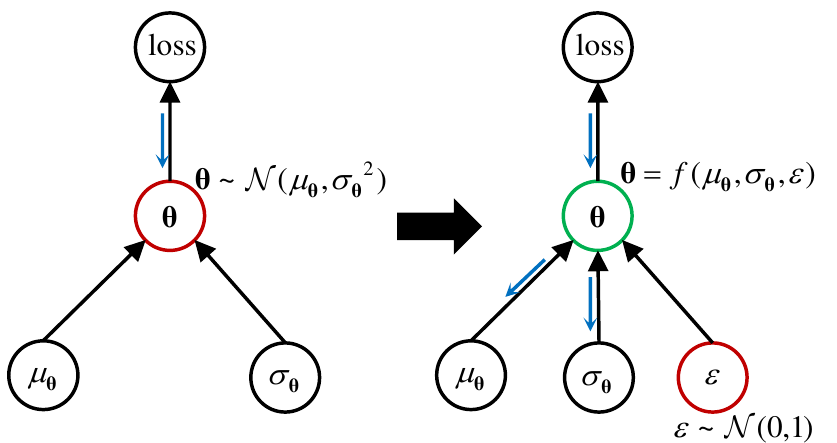}
    \caption{Graphical representation of the reparametrization trick, where by reformulating stochastic network parameters $\boldsymbol{\theta}$ as a function of statistical distribution parameters and additional stochastic inputs, the back-propagation of the loss with respect to variational parameters can be effectively computed.}
    \label{fig:Reparametrization}
\end{figure}

When a BNN is being trained, the loss function cannot be  back-propagated through $\boldsymbol{\theta}$ since it follows a probability distribution. In this scenario, the derivative of the loss with respect to the variational parameters cannot be obtained. However, the following reparameterization trick enables the formulation of $\boldsymbol{\theta}$ as a deterministic function of the variational parameters: 
\begin{equation}
    \boldsymbol{\theta} =  \mu_{\boldsymbol{\theta}}+\sigma_{\boldsymbol{\theta}} \cdot \varepsilon , \text{ where } \varepsilon \sim \mathcal{N}(0,1).
    \label{eq:reparametrization}
\end{equation}
Through this formulation, one can compute the derivative of the loss with respect to the variational parameters, as shown in Figure \ref{fig:Reparametrization}. During forward prediction runs, $\boldsymbol{\theta}$ is obtained through sampling from a standard normal distribution, $\varepsilon$, instead of sampling directly from the variational distribution $q_{\lambda}(\boldsymbol{\theta})$ so as to facilitate the implementation of the aforementioned reparametrization formulation. There are also other possible solutions for computing the gradient when random variables are included in the neural network, e.g., score function estimator,\cite{williams1992} VarGrad,\cite{Richter2020} straight-through estimator,\cite{Bengio2013} among others. The reparametrization approach described before is widely adopted in practice owing to its capability for generating unbiased gradient estimates.

It is also worth-noting that, to reduce computation efforts, the negative log-likelihood can be evaluated, in some cases, for only one $\boldsymbol{\theta}$ sample, instead of computing the expectation of several realizations, as described in Equation (\ref{eq:Loss}). The resulting gradient descent is noisy, yet it can still find its path toward the minimum loss. At the expense of drastically increasing the computational demand, a more accurate gradient can be computed by sampling multiple realizations. 

\section{Farm-wide virtual load monitoring through Bayesian neural networks} \label{sec: methodology}

\begin{figure*}[h]
    \centering    \includegraphics{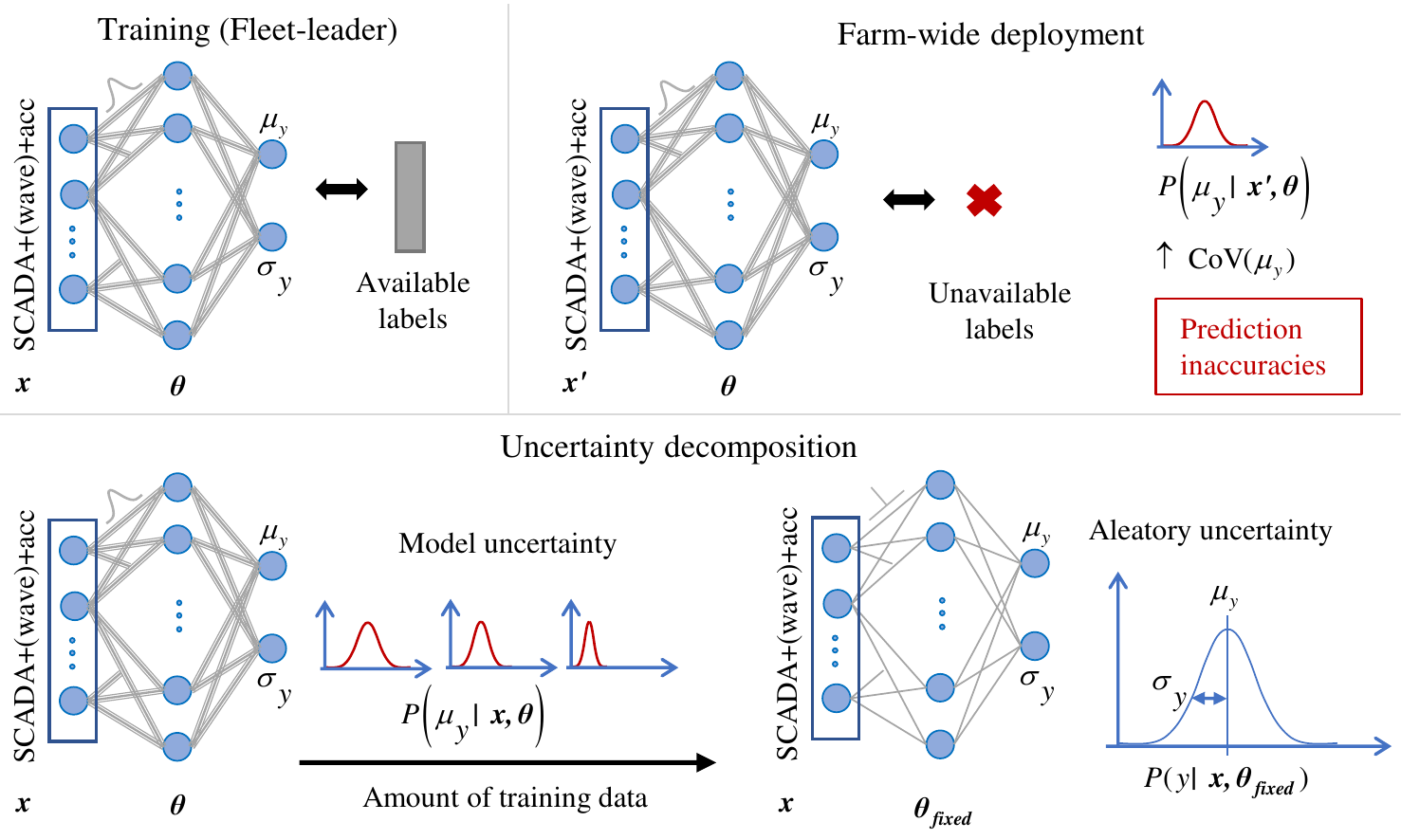}
    \caption{Rationale of the proposed farm-wide virtual load monitoring framework featuring Bayesian neural networks as data-based virtual sensors. (Top left) A fleet leader BNN is trained based on available load measurement labels. (Top right) At the deployment stage (measurement labels are no longer available), the pre-trained BNN indicates whether the generated predictions might be inaccurate by reporting a high model uncertainty. (Bottom) Uncertainty decomposition is enabled by the proposed BNN approach, yielding information on: (i) the need to collect more data for improving the model’s performance, (ii) the intrinsic variability of the analyzed phenomena.}
    \label{fig:BNN4virtualMonit}
\end{figure*}

\begin{figure}[h]
    \centering
    \includegraphics{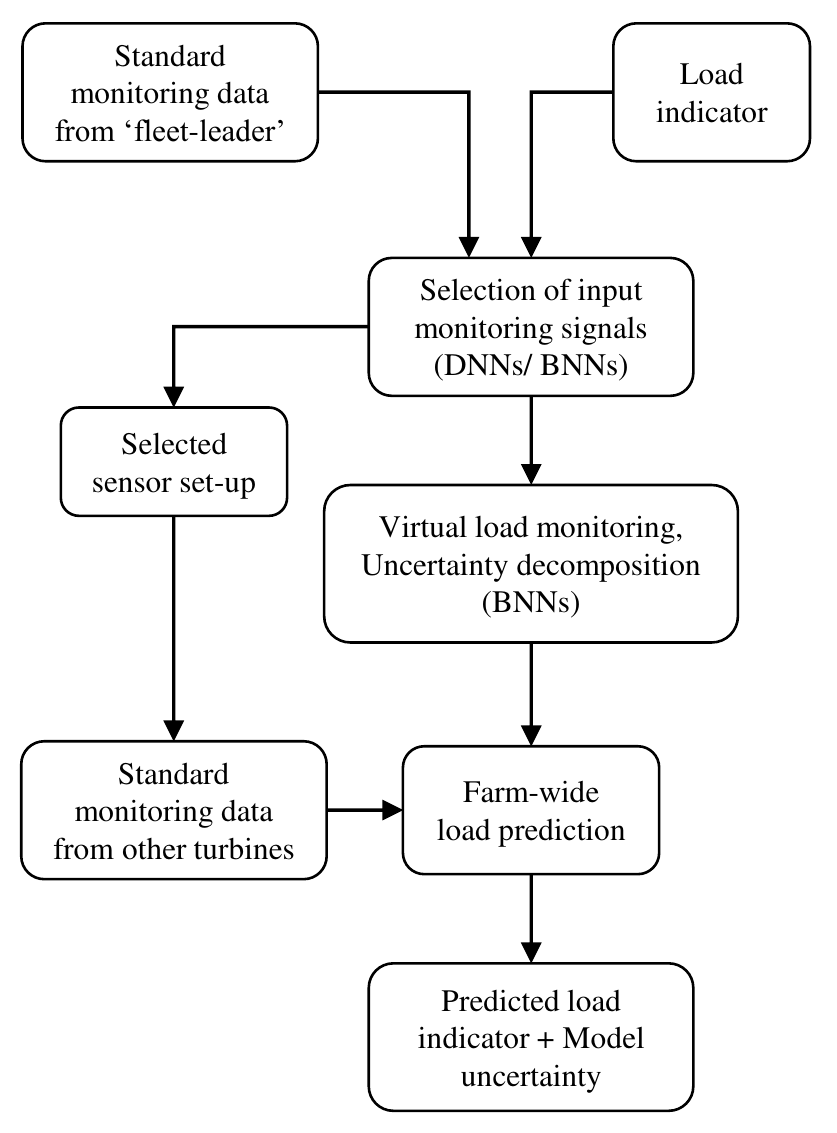}
    \caption{Flowchart diagram illustrating the steps needed for the implementation of the proposed farm-wide virtual load monitoring framework.}
    \label{fig:Method}
\end{figure}

In general, the full instrumentation of an offshore wind farm with strain sensors becomes economically impractical due to the elevated installation and maintenance costs associated with the process. In this context, virtual load monitoring offers an efficient alternative by providing load information - denoted hereafter as `load indicator' - based on readily available monitoring data - denoted hereafter as `standard monitoring data', e.g., SCADA and accelerometer data. Thus, virtual load monitoring constitutes a natural solution for the implementation of a farm-wide monitoring strategy, i.e., one or a set of representative turbines, commonly designated as a fleet-leader, is fully instrumented enabling the training of a data-based model, which then provides load indicator predictions to the other non-fully instrumented wind turbines. As opposed to conventional deterministic virtual load monitoring schemes, this paper proposes a probabilistic virtual monitoring method, which by indicating the uncertainty associated with the `load indicator', intrinsically informs the quality of the generated predictions. The overarching rationale of the proposed virtual monitoring method is illustrated in Figure \ref{fig:BNN4virtualMonit}, highlighting BNNs' capabilities for automatically detecting potential prediction inaccuracies when the virtual model is deployed to wind turbines where load measurements are not available. Moreover, it is also showcased in the figure how the overall involved uncertainty can be decomposed into model and aleatory components, providing information on (i) whether more data is needed to improve model's performance, and (ii) capturing the intrinsic variability associated with the analyzed phenomena, respectively. A more specific description of the general procedure for implementing a BNN-based virtual load monitoring model is summarized in Figure \ref{fig:Method}, and will be further explained in the following subsections.

\subsection{Selection of the input monitoring signals}

In principle, various environmental, operational, and/or motion signals, e.g., SCADA, wave data, accelerations, can be monitored and fed as inputs to the virtual load monitoring model. However, a reduced selection of meaningful monitoring signals as inputs to the model will ease the instrumentation setup of non-fully monitored wind turbines, and overall alleviate practical constraints during the deployment of the farm-wide virtual monitoring strategy. In order to select the reduced set of monitoring signals, a data-based model can be tested for various potential configurations, and by observing the generalization error associated with each considered setting, the setup that results in the desirable trade-off between accuracy and monitoring equipment complexity can be then chosen. 

The generalization error can be estimated through either deterministic or probabilistic approaches, yet the metrics employed to assess the error vary between them. 
Whereas deterministic neural networks (DNNs) are constructed assuming point estimates for the constituent weights and biases, Bayesian neural networks (BNN) consider weights and biases as probability distributions, as shown in Figure \ref{fig:DNN_BNN}. Besides, even if both DNNs and BNNs similarly receive input monitoring signals $\boldsymbol{x}= \{x_1, x_2, ... x_M\}$, DNNs provide a deterministic load indicator as output, while BNNs’ output becomes a probabilistic load indicator. For the case of deterministic neural networks, the standard metrics, e.g., mean absolute error (MAE) or root mean squared error (RMSE), can be used as reference to compare the performance of the analyzed virtual load monitoring models, as:
\begin{equation}
    MAE = \frac{1}{N}\sum_{i=1}^N{\left|y_i-\hat{y}_i\right|},
\end{equation}

\begin{equation}
    RMSE = \sqrt{\frac{1}{N}\sum_{i=1}^N{\left(y_i-\hat{y}_i\right)^2}},
\end{equation}
where $N$ stands for the number of test samples, and $\hat{y}$ represents the model prediction, compared against the ground truth (label), $y$. 

On the other hand, a probabilistic output is provided by BNNs, as mentioned previously, from which random samples can be drawn. More specifically, the output layer features the statistical parameters of a specified probability distribution, e.g., a Gaussian $y \sim \mathcal{N}(\mu_y,\sigma_y)$, which are then fed to an additional distribution layer to be able to draw random samples of the load indicator $\hat{y}$, as well as to compute likelihood of the label, i.e., $P(y\mid \boldsymbol{x}, \boldsymbol{\theta})$.  To assess the performance of a BNN, one cannot rely on MAE or RMSE, since the model outputs are random realizations. Instead, metrics that provide a probabilistic interpretation should be observed, e.g., the expected log-likelihood of the label given the prediction model, defined as:
\begin{equation}
    \mathbb{E}[\mathcal{L}(y)]= \frac{1}{N}\frac{1}{N_f}\sum_{i=1}^N\sum_{N_f}\log {(P(y_i\mid \mu_{y,i},\sigma_{y,i}))},
    \label{eq:expected_loglikelihood}
\end{equation}
where $\mu_y$ and $\sigma_y$ stand for the output statistical parameters predicted by the model. Note that the output statistical parameters are, for the case of BNNs, also stochastic, resulting from the random realizations drawn from the network’s stochastic weights and biases, and thus statistical properties of the likelihood can be retrieved via numerical simulations, i.e., sampling $N_f$ forward predictions.

\subsection{Bayesian neural networks uncertainty decomposition}
As explained before, the statistics of BNN’s predicted results can be computed at the deployment stage through numerical simulations, e.g., Monte Carlo sampling. For instance, one can estimate the expected value and the predictive uncertainty of the load indicator $\hat{y}$ given newly acquired standard monitoring data $\boldsymbol{x}$:
\begin{equation}
    \mathbb{E}[\hat{y}\mid \boldsymbol{x}] = \frac{1}{N_f}\sum_{N_f}{f(\hat{y}\mid \boldsymbol{x},\boldsymbol{\theta})},
    \label{eq:expectation}
\end{equation}

\begin{equation}
    \mathbb{V}(\hat{y}\mid \boldsymbol{x}) = \frac{1}{N_f}\sum_{N_f}\left({f(\hat{y}\mid \boldsymbol{x},\boldsymbol{\theta})}\right)^2-\left(\mathbb{E}[\hat{y}\mid \boldsymbol{x}\right])^2,
    \label{eq:Predictive_uncertainty}
\end{equation}
where the network parameters, $\boldsymbol{\theta}$ are randomly drawn from the posterior weights and biases associated distributions, and $f$ symbolizes the Bayesian network model itself. Note that, in this case, the retrieved predictive uncertainty estimate $\mathbb{V}(\hat{y}\mid \boldsymbol{x})$ encompasses both aleatory and epistemic contributions. 

On the one hand, aleatory uncertainties arise from the inherent randomness of physical phenomena and/or the presence of noise in sensing devices. While the physical uncertainty is irreducible, \cite{Depeweg2018b} measurement uncertainty can be reduced by modulating the noise level of sensors, albeit it cannot be controlled by adjusting the model. On the other hand, epistemic uncertainties are induced by the quality of the model and can be reduced by improving the model. For instance, at the deployment stage, a trained BNN might indicate high epistemic uncertainty if data outside of the training domain is fed to the network, and after the model is retrained from representative data in the reported high uncertainty region, the BNN’s epistemic uncertainty can be further reduced. Theoretically, the epistemic uncertainty will be totally dissipated in the limit of infinite available training data. In practice, however, there exists no perfect model for predicting the response of complex engineering systems, i.e., the model might not consider all representative features, and the additional uncertainty associated with the missing or unavailable latent variables is sometimes also denoted as aleatory uncertainty. \cite{Depeweg2018a} 
 
A decomposition of the overall uncertainty retrieved by the BNNs, into its aleatory and epistemic contributions, becomes highly informative when deploying the trained network to the farm-wide level. An indication of high global predictive uncertainty does not inherently report the need for retraining the model, since the variability might correspond to noise present in the observations (labels). However, further data collection and model retraining actions can be planned as a result of observed high model uncertainty. Not isolated to virtual load monitoring applications, uncertainty decomposition is an active topic within the probabilistic machine learning community.\cite{Depeweg2018a, Depeweg2018b, Wang2020}
The overall uncertainty can be decomposed, according to the law of total variance, as follows:
%, and thus the model (epistemic) uncertainty can be observed in the variance of the output distribution parameters:
\begin{equation}
    \mathbb{V}(\hat{y}\mid \boldsymbol{x}) = \mathbb{E}[\sigma_y^2\mid \boldsymbol{x}]+\mathbb{V}(\mu_y\mid \boldsymbol{x})
    \label{eq:Toal_variance}.
\end{equation}
In general, complex engineering systems are exposed to aleatoric physical phenomena, i.e., the system response for a given set of input parameter values
 does not correspond to a single output value. Since conventional numerical simulators are very often deterministic, the intrinsic variability is normally accounted for by running multiple simulations with different random seeds for each input combination, e.g., wind and wave conditions specified following offshore wind design practices and recommendations. \cite{dnvj101, iec614003} In BNNs, the aleatory uncertainty is captured by learning the variance parameter $\sigma_y^2$, thus inherently yielding a probabilistic output. The first term $\mathbb{E}[\sigma_y^2\mid \boldsymbol{x}]$ in Equation (\ref{eq:Toal_variance}) can be interpreted as the aleatory component, computed as:
\begin{equation}
    \mathbb{E}[\sigma_y^2\mid \boldsymbol{x}] = \frac{1}{N_f}\sum_{N_f}({f(\sigma_y\mid \boldsymbol{x},\boldsymbol{\theta})})^2,
\end{equation}
The epistemic uncertainty is encapsulated in the probability distribution of the network's weights and biases. The variance of the BNN's predicted means $\mathbb{V}(\mu_y\mid \boldsymbol{x})$, computed as in the following equation, therefore explains the epistemic uncertainty,
\begin{equation}
    \mathbb{V}(\mu_y\mid \boldsymbol{x}) = \frac{1}{N_f}\sum_{N_f}{\left(f(\mu_y\mid \boldsymbol{x},\boldsymbol{\theta})\right)^2}-\left(\mathbb{E}[\mu_y\mid \boldsymbol{x}]\right)^2.
    \label{eq:model_uncertainty}
\end{equation}
where,
\begin{equation}
    \mathbb{E}[\mu_y\mid \boldsymbol{x}] = \frac{1}{N_f}\sum_{N_f}{f(\mu_y\mid \boldsymbol{x},\boldsymbol{\theta})},
    \label{eq:Expectation_mu}
\end{equation}
It should be noted that there also exists the uncertainty of BNN's predicted variance $\mathbb{V}(\sigma_y^2\mid \boldsymbol{x})$, yet its contribution to the overall uncertainty is insignificant enough to be neglected.

% \begin{equation}
%     \mathbb{V}(\sigma_y\mid \boldsymbol{x}) = \frac{1}{N_f}\sum_{N_f}\left({f(\sigma_y\mid \boldsymbol{x},\boldsymbol{\theta})}-\mathbb{E}(\sigma_y\mid \boldsymbol{x})\right)^2.
%     \label{eq:model_uncertainty}
% \end{equation}
% The output parameters can also be combined into one variable such that $cv_y = {\sigma_y}/{\mu_y}$, and the model uncertainty therefore becomes:
% \begin{equation}
%     \mathbb{V}(cv_y\mid \boldsymbol{x}) = \frac{1}{N_f}\sum_{N_f}\left({f(cv_y\mid \boldsymbol{x},\boldsymbol{\theta})}-\mathbb{E}(cv_y\mid \boldsymbol{x})\right)^2.
%     \label{eq:model_uncertainty}
% \end{equation}

\subsection{Farm-wide load prediction}
Once the reduced set of input monitoring signals, i.e., standard monitoring data, has been identified by quantifying the generalization error through either deterministic or Bayesian neural networks, and the BNN model corresponding to the fleet-leader has also been trained, the following steps are to be implemented for farm-wide load prediction:
\begin{itemize}
\item Collection and treatment of the required standard monitoring data $\boldsymbol{x}$ from non-fully monitored turbines.
\item Deployment of the trained BNN on non-fully monitored turbines, from which multiple forward simulations are run, thus randomly drawing load indicator realizations, $\hat{y}$.
\item Computation of the load indicator expected value $\mathbb{E}[\hat{y}\mid \boldsymbol{x}]$.
%its predictive uncertainty $\mathbb{V}(\hat{y}\mid \boldsymbol{x})$.
\item  Estimation of the model epistemic uncertainty $\mathbb{V}(\mu_y\mid \boldsymbol{x})$, and the performance metric $\mathbb{E}[\mathcal{L}(y)]$ if the target labels are available, and further information collection actions might be decided depending on the observed model uncertainty metric.
\end{itemize}

\subsection{Epistemic Bayesian neural network}

In certain cases, one might only be interested in a virtual sensor model that provides a mapping between the inputs and deterministic predicted output(s). In that case, an epistemic variant of the proposed BNN can be implemented by assigning zero to the output variance node(s) (i.e., $\sigma_y \to 0$), hence disregarding the potential aleatory uncertainty associated with the output response and only seeking the prediction of the mean response output node (i.e., $\mu_y$). Theoretically, the response of typical engineering systems normally contains physical aleatory uncertainty, yet more data is required to train a probabilistic output response compared to a point estimate. In any case, an epistemic BNN can yield model uncertainty information because the weights and biases are still described by probability distributions. Since no aleatory uncertainty is now incorporated into the model, i.e., the first term in Equation (\ref{eq:Toal_variance}) becomes zero and the total predictive uncertainty directly corresponds to model uncertainty. Whether aleatory uncertainty is included or not, farm-wide deployment can be executed in both cases since the BNN model will inform valuable model uncertainty metrics about its confidence in the generated predictions.

\section{Experimental campaign: Probabilistic virtual load monitoring in an offshore wind farm}
The proposed framework for farm-wide virtual load monitoring is hereafter implemented and tested for the specific case of an existing offshore wind farm. In this study, we do not rely only on environmental and operational data, but also incorporate acceleration signals within the standard monitoring data that will be used to predict the load indicator, as motion information provides an indication of wind turbines’ structural dynamics and significantly influences fatigue load estimations.\cite{Santos2022} 

\begin{figure}
    \centering    \includegraphics{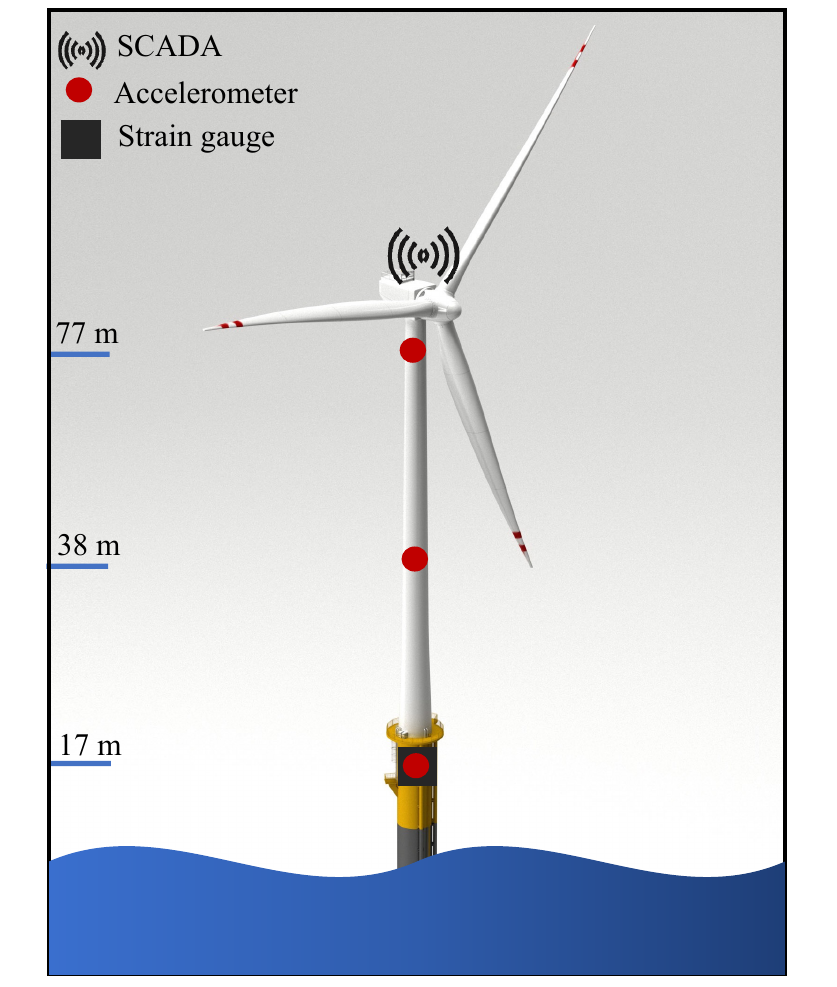}
    \caption{Illustration depicting the monitoring setup installed on an operational offshore wind turbine, from which data was continuously collected during the course of the experimental campaign. The monitoring setup includes a standard SCADA system, accelerometers at three different levels, and strain gauges installed at the lowest level.}
    \label{fig:SensorSetup}
\end{figure}

\subsection{Monitoring setup and dataset description}
This investigation relies on the data collected during the course of a 2-year monitoring campaign, from early 2018 to the end of 2019, conducted on three wind monopile-supported turbines located in a Belgian offshore wind farm. The overall monitoring setup installed above the mean sea level is graphically illustrated in Figure \ref{fig:SensorSetup}. The turbines are equipped with a standard supervisory, control and data acquisition (SCADA) system, continuously collecting both environmental and operational data. 
More specifically, the collected SCADA data contains 10-minute statistics of wind speed and direction, turbine rotational speed, yaw and pitch attitude, and instantaneous power. The sensor setup does not feature wave measurements, yet data from Meetnet Vlaamse Banken\cite{AgentMDK}’s Westhinder wave buoy was additionally collected, thus providing wave height, wave period and wave direction information. As aforementioned, the monitored turbines are also equipped with three accelerometers installed on the transition piece and tower at 17 m, 38 m, and 77 m above lowest astronomical tide (LAT), respectively. The collected accelerometer data reports acceleration statistics, i.e., minimum, maximum, and root-mean-square (rms), from 10-minute time series of fore-aft (FA) and side-to-side (SS) accelerations.

\begin{table*}
\small\sf\centering
\caption{Description of the dataset.}
\begin{tabular}{lllll}
\toprule
&Sensor& Monitoring signal & Symbol & Units\\
\midrule
\multirow{16}{*}{\rotatebox[origin=c]{90}{Input}} &SCADA & Rotational speed (mean) & $\mu[RPM]$ & rpm\\
& & Yaw angle (mean) & $\mu[Yaw]$ &deg\\
& & Pitch angle (mean) & $\mu[Pitch]$ & deg\\
& & Power (mean) & $\mu[Power]$
& kW\\
& & Wind speed (mean) & $\mu[WSpd]$ & m/s\\
& & Wind speed (std) & $\sigma[WSpd]$ &m/s\\
& & Wind direction (mean)& $\mu[WDir]$ & deg\\
\cmidrule{2-5}
& Wave buoy & Wave height &$H_s$ & cm\\
& & Average wave period &$T_p$ & s\\
& & Wave direction & $\theta_{w}$ & deg\\
\cmidrule{2-5}
&  Accelerometers & FA acceleration (max) & $max[acc_{FA}]$& g\\
&  - LAT-017 & FA acceleration (min) & $min[acc_{FA}]$& g\\
&  - LAT-038 & FA acceleration (rms) & $rms[acc_{FA}]$& g\\
&  - LAT-077 & SS acceleration (max) & $max[acc_{SS}]$& g\\
&  & SS acceleration (min) & $min[acc_{SS}]$& g\\
&  & SS acceleration (rms) & $rms[acc_{SS}]$& g\\
\midrule
\multirow{3}{*}{\rotatebox[origin=c]{90}{Output}} &Strain gauges & DEM (side-to-side) & $DEM_{tl}$ & MNm\\
& - LAT-017 & DEM (fore-aft) & $DEM_{tn}$ & MNm\\
\\
% \midrule
% \multicolumn{3}{l}{Number of 10-minute files (Fleet-leader): 70404}\\
% \multicolumn{3}{l}{Number of 10-minute files (MP01): 37820}\\
% \multicolumn{3}{l}{Number of 10-minute files (MP02): 46319}\\
\bottomrule
\end{tabular}\\
\label{tab:Description of the dataset}
\end{table*}

Along with the standard monitoring data mentioned previously, strain gauges are installed at 17 m above the lowest astronomical tide (LAT) level of the wind turbines, so as to collect load signals, i.e., bi-axial bending moments, $M_{tl}$ in SS direction and $M_{tn}$ in FA direction. Following common offshore wind industrial and scientific practices,\cite{iec614003} the time series of the monitored loads are post-processed into representative damage equivalent loads (DELs), i.e., equivalent load range such that the damage caused by a pre-defined number of the equivalent load amplitude cycles $N_{eq}$ equals the damage $D_{SN}$ caused by the original load time series, as computed using Miner-Palmgren's rule:
\begin{equation}
    D_{SN} = \sum_{i=1}^{N_s}{\frac{n_{i}}{k\cdot S_{r,i}^{-m}}}= \frac{N_{eq}}{k\cdot DEL^{-m}},
\end{equation}
where $k$ and $m$ correspond to linear S-N curve parameters, $N_s$ stands for the number of stress range bins in the load spectrum, whereas $S_{r,i}$ and $n_i$ represent the reference value for the $i^{th}$ stress range bin and the number of cycles inside that bin, respectively. Note that the equivalent stress cycles $N_{eq}$ are commonly specified as $10^7$. In this work, the damage equivalent moment (DEM) is computed for each retrieved 10-minute time series of bending moment measurements, as follows:
\begin{equation}
    DEM = \left(\frac{\sum_{i=1}^{N_m}{n_i\cdot M_{r,i}^{m}}}{N_{eq}}\right)^{1/m},
\end{equation}
where $N_m$ stands for the number of bins in the load spectrum, $M_{r,i}$ and $n_i$ represent the reference moment value for the $i^{th}$ bin and the number of cycles in that bin. One can straightforwardly compute the fatigue damage from the calculated DEM by also considering the geometrical properties (i.e., thickness, second moment of area), as well as the specified SN curve parameters $k$ and $m$. In this application, the estimated damage equivalent moments $DEM_{tl}$ and $DEM_{tn}$ constitute the output load indicators that are provided as labels to the virtual load monitoring model.

In summary, Table \ref{tab:Description of the dataset} lists and describes each dataset considered in this work, overall containing 28 monitoring input signals: 7 from the SCADA system, 3 from the wave buoy, and 18 from the accelerometers; and 2 monitoring output signals retrieved from the strain gauges. As aforementioned, the data is recorded for three wind turbines within the same offshore wind farm, and during the investigation, the data retrieved from the fleet-leader wind turbine is employed for the development and training of the virtual load monitoring model, where the data is randomly split into 75\% for training and 25\% for testing. The data collected from the other two turbines is fully reserved for farm-wide load prediction purposes.
% \begin{figure}[h][h]
%     \includegraphics{Figures/Method1.pdf}
%     \caption{Framework of development and deployment of probabilistic virtual load monitoring model.}
%     \label{fig:Method1}
% \end{figure} 

\subsection{Selection of the input monitoring signals}
 
Following the procedures provided in the framework, a reduced set of representative and informative input monitoring data is carefully selected through both deterministic and Bayesian neural network approaches. In order to decide the reduced set of standard monitoring data, the generalization error is computed for the following model configurations of monitoring input signals:
\begin{enumerate}
    \item SCADA + wave
    \item SCADA + wave + accelerometer (LAT-017)
    \item SCADA + wave + accelerometer (LAT-038)
    \item SCADA + wave + accelerometer (LAT-077)
    \item SCADA + wave + accelerometers (LAT-017, 038)
    \item SCADA + wave + accelerometers (LAT-017, 038, 077)
    \item [7.]SCADA
    \item [8.]SCADA + accelerometer (LAT-017)
    \item [9.]SCADA + accelerometer (LAT-038)
    \item [10.]SCADA + accelerometer (LAT-077)
    \item [11.]SCADA + accelerometers (LAT-017, 038)
    \item [12.]SCADA + accelerometers (LAT-017, 038, 077)
\end{enumerate}
Note that a distinction has been made in the selection process between model configurations which include, or do not include, wave data. It is worth exploring alternatives that purely use the turbine's monitoring data without relying on the secondary wave data, thus naturally simplifying the later implementation of a farm-wide load monitoring. 

\begin{figure*}
     \centering
     \begin{subfigure}[b]{\textwidth}
         \centering         \includegraphics{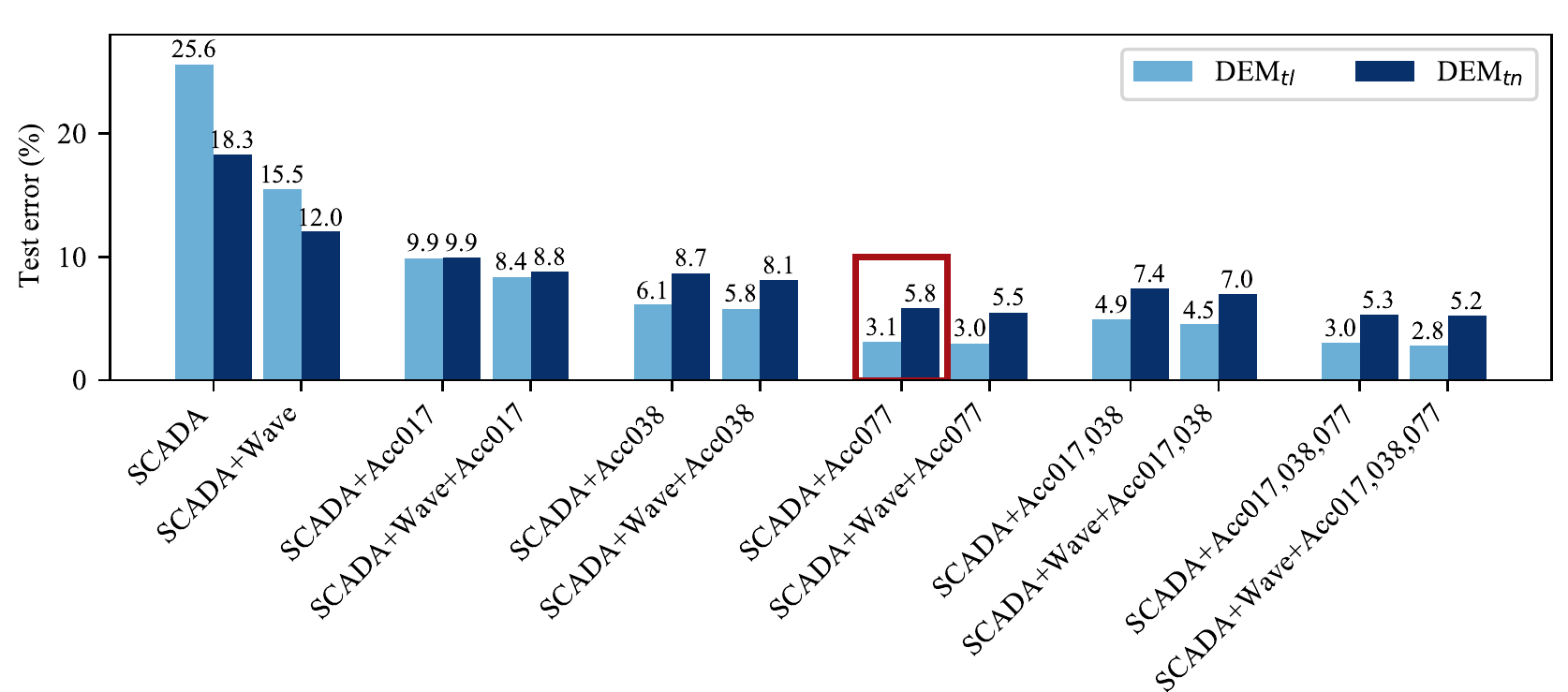}
         \caption{Errors reported during the testing stage of the implemented deterministic neural networks. The lower error indicates a better performance of the model.}
         \label{fig:DNNErrors_merge}
     \end{subfigure}
     \vfill
     \begin{subfigure}[b]{\textwidth}
         \centering         \includegraphics{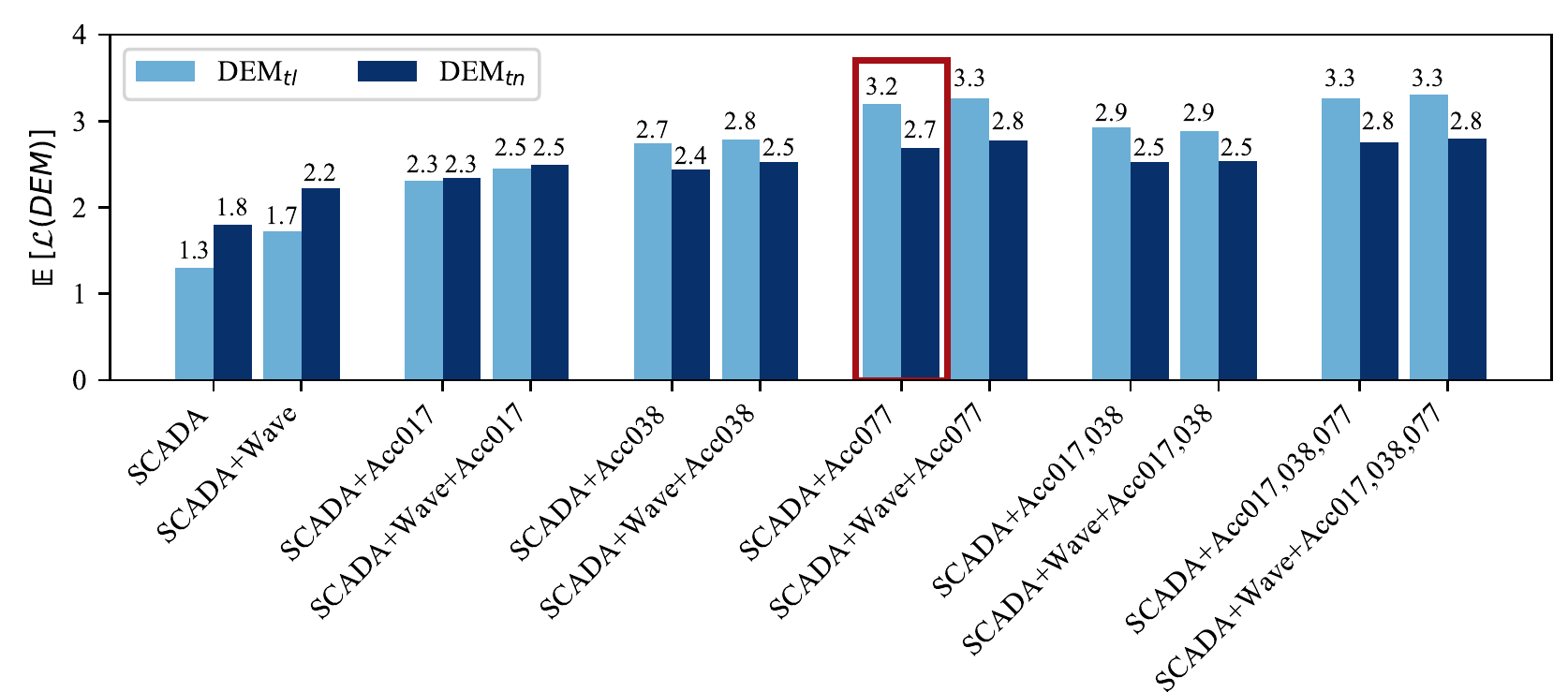}
         \caption{Expected log-likelihood reported during the testing stage of the implemented Bayesian neural networks. The higher expected log-likelihood indicates a better performance of the model.}
         \label{fig:BNNErrors_merge}
     \end{subfigure}
     \caption{Illustration showcasing the performance of load prediction data-based models. Each bar corresponds to a model specified with a specific set of input monitoring signals, i.e., SCADA, wave, and/or acceleration data. The red box indicates the selected reduced set of input monitoring signals.}
     \label{fig:Errors}
\end{figure*}

\subsection{Deterministic neural networks}
The deterministic neural networks (DNNs) implemented in this investigation rely on \textit{Keras}\cite{chollet2015} API, which forms part of the machine learning and artificial intelligence library \textsc{TensorFlow}. The topology of the DNNs features a fully connected feed-forward neural network consisting of three hidden layers with 64, 128, 64 neurons, respectively. Naturally, the width of the input layer varies according to the tested input monitoring signal combination, and the output layer features two neurons corresponding to the predictions of damage equivalent moments in the side-to-side $(DEM_{tl})$ and fore-aft $(DEM_{tn})$ directions. Note that all network layers are equipped with rectified linear unit (\textit{ReLU}) activation functions.  

During the course of the training task, the neural networks are trained via \textit{Adamax} optimizer at the default learning rate of 0.001, minimizing the loss function corresponding to the mean absolute error (MAE):
\begin{multline}
    MAE = \frac{1}{2N}\sum_{i=1}^N\Bigg(\left|DEM_{tl,i}-\hat{DEM}_{tl,i}\right|
    \\+\left|DEM_{tn,i}-\hat{DEM}_{tn,i}\right|\Bigg),
\end{multline}
% \begin{equation}
%     MAE = \frac{1}{2N}\sum_{i=1}^N\Bigg(\left|DEM_{tl,i}-\hat{DEM}_{tl,i}\right|
%     +\left|DEM_{tn,i}-\hat{DEM}_{tn,i}\right|\Bigg),
% \end{equation}
where $N$ stands for the total number of training samples, $DEM$ and $\hat{DEM}$ represent the ground truth (measurements) and predicted damage equivalent moments, respectively. For each tested combination, the training task conducted 200 epochs, running stochastic gradient descent based on randomly collected 32-sample batches. Moreover, potential overfitting conflicts are avoided by implementing an `early stopping callback’, i.e., the training task is stopped if there is no improvement in the validation MAE. \cite{Hlaing2022} In this regard, the training dataset is randomly split into 80\% for training and 20\% for validation purposes.

As described before, we have overall tested 12 input monitoring signal setups. Figure 
 \ref{fig:DNNErrors_merge} illustrates the performance of each DNN model with respect to the considered input monitoring setup, indicating in  the generalization error computed as: 
\begin{equation}
    Error (\%) = \frac{100}{N}\sum_{i=1}^N{\left|\frac{DEM_i-\hat{DEM}_{i}}{DEM_{i}}\right|},
    \label{eq:percentage_error}
\end{equation}
where $N$ stands for the total number of test samples. Interestingly, including wave data as an input to the model is beneficial in configurations in which information on accelerations is not available, where one can observe a reduction of approximately 6.3\% $(DEM_{tn})$ and 10.1\% $(DEM_{tl})$ in the computed generalization error. However, the benefit of feeding the network with wave data becomes negligible once acceleration information becomes also available, resulting in around $1\%$ reduction of the generalization error in all accelerometer-integrated setups. As a result, only primary wind turbine monitoring signals, i.e., combination of SCADA system and accelerometers, can be deemed, in this case, as a satisfactory input monitoring setup, thus avoiding the need of relying on secondary wave data during the deployment stage of the virtual load monitoring model.

Furthermore, it can also be observed that the generalization error of the model including SCADA + accelerometer (LAT-077) as inputs remains very similar to the error reported for the input setup that features SCADA + accelerometers (LAT-017, 038, 077). Specifically, the generalization error for the case in which all accelerometers information is considered results in 3.0\% $(DEM_{tl})$ and 5.3\% $(DEM_{tn})$, whereas the error corresponding to the setup where only the top accelerometer is included as input to the model results in 3.1\% $(DEM_{tl})$ and 5.8\% $(DEM_{tn})$, respectively. Installing two additional accelerometers at the lower levels reduces only 0.1-0.5\% in the errors. 
Since fatigue is primarily driven by the first structural mode to which the top accelerometer is more sensitive compared to the other two lower levels, it provides the most informative data to predict the loads.
Therefore, the input monitoring signal setup SCADA + top accelerometer interestingly outperforms the monitoring input combination of SCADA + accelerometers (LAT-017, 038), even though the load prediction is conducted for a location situated at the lowest level (LAT-017).

\subsection{Bayesian neural networks}

The Bayesian neural networks introduced in this investigation are implemented with the support of the probabilistic deep learning library \textsc{TensorFlow Probability}. As for the case of the DNNs, the width of the Bayesian neural network (BNN) input layer is defined based on the specified input monitoring data, along with three hidden layers equipped with \textit{ReLU} activation functions, and an output layer with 4 neurons, from which the output statistical parameters $\mu_{DEM_{tl}}, \sigma_{DEM_{tl}}, \mu_{DEM_{tn}}, \sigma_{DEM_{tn}}$ are estimated. Note that, for the case in which the standard deviation of the output load indicator distribution is known and fixed, the BNN can be alternatively laid out with a reduced output layer containing only the mean statistical parameter that drives the resulting load distribution. 
More specifically, both hidden and output layers of the BNNs are built through \textit{DenseFlipout}, implementing Bayesian variational inference via a Flipout estimator. Since the training samples in each batch share the stochastic weights $\varepsilon$, there is potential correlation in the resulting gradients which can lead to inefficient training. Flipout is an efficient method to improve variance reduction by implicitly sampling pseudo-independent stochastic weights for each training data and therefore decorrelating the gradients within a batch. For more detailed description of the Flipout estimator, the reader is referred to Wen et al.\cite{Wen2018}

\begin{figure}
    \centering
    \includegraphics{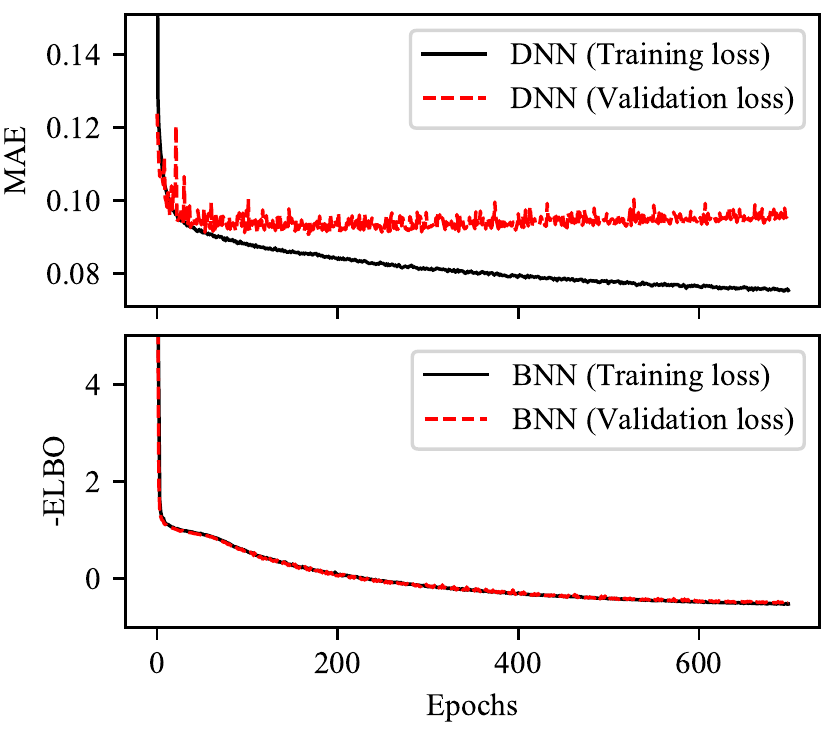}
    \caption{Graphical representation and comparison between the usual training behavior of [top] standard deterministic neural networks (DNNs) and [bottom] Bayesian neural networks (BNNs). Training and testing losses are plotted for both models over epochs. The automatic overfitting control featured by BNNs can be observed in the illustration. }
    \label{fig:Overfit}
\end{figure}

A thorough description of the topology and training environment considered for both deterministic and Bayesian neural networks is showcased in Appendix \hyperlink{AppendixA}{A}, providing details for each tested input monitoring setup. In general, the number of neurons included in the hidden layers of BNNs is lower than the neurons specified for the deterministic counterparts. If the same architecture would be specified for both DNNs and BNNs, the training for the latter will naturally take longer as Bayesian inference normally demands more computational resources than classical backpropagation. However by considering the network parameters as probability distributions, BNNs are intrinsically more informative, despite having less neurons, and additionally provide beneficial properties to avoid overfitting with respect to standard DNNs. As shown in Figure \ref{fig:Overfit}, a DNN starts to overfit the training data after some training episodes, i.e., the validation loss plateaus while the training loss keeps decreasing. One therefore needs to add regularizers and/or test on a validation set after each training epoch. In this study, an early stopping callback based on validation loss metrics is implemented in order to prevent potential overfitting. On contrary, a BNN does not overfit as it contains a regularizer by default  as explained in the previous section, and a separate validation dataset is no longer needed, and the early stopping callback in BNN monitors the training loss.

\begin{figure*}
     \centering
     \begin{subfigure}[b]{0.49\textwidth}
         \centering
         \includegraphics{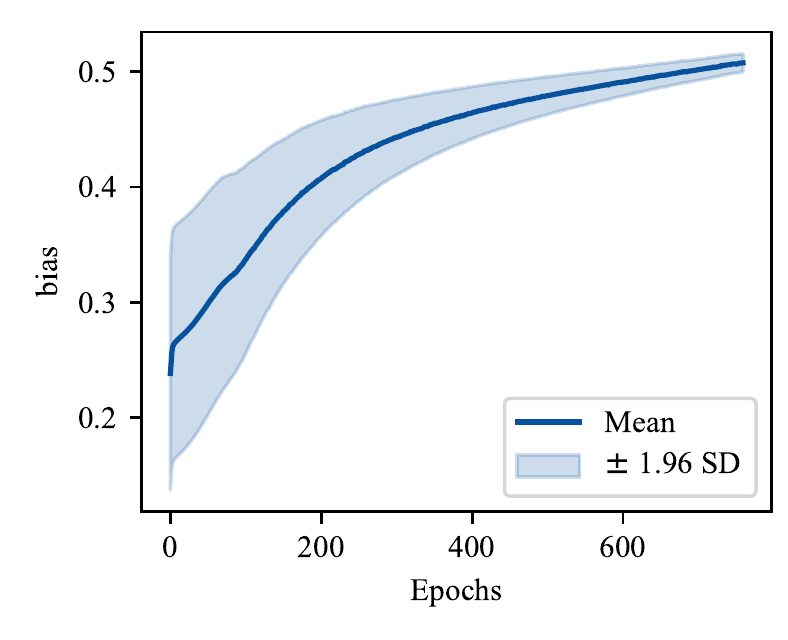}
         \caption{Illustration representing the bias' mean with a continuous dark blue line and confidence intervals ($\pm$ 1.96 SD) with a light blue color shade.\\{~}}
     \end{subfigure}
     \hfill
     \begin{subfigure}[b]{0.49\textwidth}
         \centering
        \includegraphics{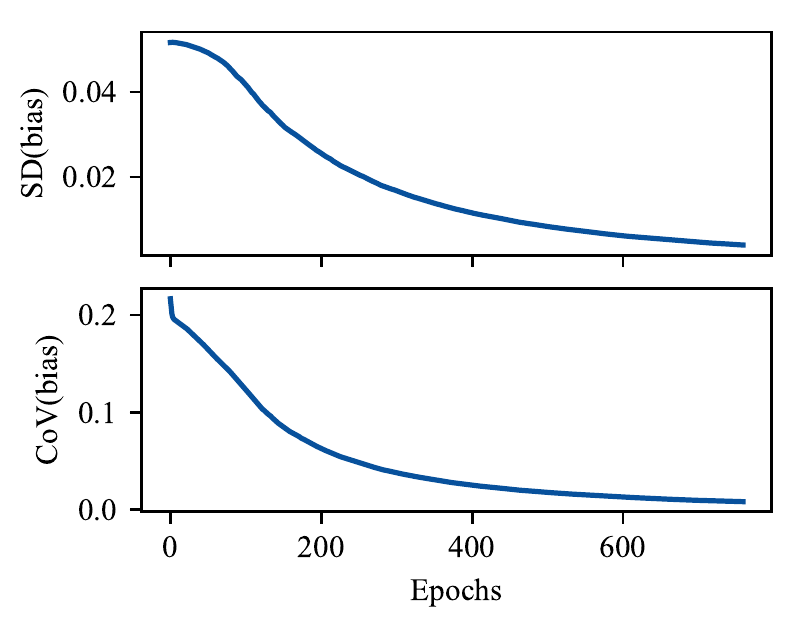}
         \caption{[Top] Bias' standard deviation, SD(bias), represented over training epochs. [Bottom] Bias' coefficient of variation, CoV(bias), plotted during the course of the training stage.}
     \vfill
     \end{subfigure}
    %  \begin{subfigure}[b]{0.49\textwidth}
    %     \centering
    %     \includegraphics{Figures/Bias3.pdf}
    %     \caption{Mean and CI ($\pm$ 1.96 Std) of the output layer bias 2.\\{~}}
    %  \end{subfigure}
    %  \begin{subfigure}[b]{0.49\textwidth}
    %     \centering
    %     \includegraphics{Figures/Bias3_cov.pdf}
    %     \caption{Standard deviation and coefficient of variation of the output layer bias 2.}
    %  \end{subfigure}
     \caption{Evolution of a specific bias from the neural network over training epochs. The reduction of model uncertainty can be appreciated by observing the plunge of the bias' coefficient of variation (CoV) over the course of the training task.}
     \label{fig:Biases}
\end{figure*}

The prior weights' distributions of the Bayesian neural networks are assigned to follow a multivariate standard normal distribution. Since the BNN needs to minimize the negative ELBO, i.e., the sum of KL divergence and negative log-likelihood as shown in Equation (\ref{eq:Loss}), the built \textit{DenseFlipout} layers add the KL divergence between the posteriors and their respective priors to the specified loss function. The Bayesian model is then trained on 1024-sample data batches applying the \textit{Adam} optimizer. On a worth-noting remark, we used a very small learning rate to train BNNs in this work since otherwise, the network tends to converge to a local minimum owing to the stochastic nature of the model combined with the complexity of the dataset. 

The training is established for 2000 epochs but the BNN early stopping criteria demands to stop if there is no significant improvement in the training loss. 
% via a state-of-the-art variational inference method
The evolution of the weights can also be tracked while the Bayesian neural network is being trained. Figure \ref{fig:Biases} illustrates the evolution of the mean, the standard deviation and the coefficient of variation of a model bias in the neural network. Evidently, rapid reduction of the uncertainty can be observed as the network learns from the dataset. 

The comparison of BNN models is shown in Figure \ref{fig:BNNErrors_merge}. The expected log-likelihood on the test dataset is computed from 10,000 forward runs according to Equation (\ref{eq:expected_loglikelihood}). Similar to the deterministic case, the model which maximizes the expected log-likelihood is the one with all available information, however, the model fed with reduced information - SCADA + accelerometer (LAT-077) - also provides satisfactory results. Furthermore, since both deterministic and probabilistic approaches provide consistent results, one can perform the sensor sensitivity analysis only in a deterministic manner, if preferred, without particular interest on model uncertainty. 

Whereas the optimization of sensor placement can be application specific, \cite{Ostachowicz2019} here we aimed to maximize the model performance, yet keeping the required instrumentation as minimal as possible to facilitate farm-wide deployment. Therefore, based on the presented results, we selected the SCADA + accelerometer (LAT-077) combination for subsequent steps. 

\subsection{Fleet-leader's virtual monitoring model}

\begin{figure*}[h]
     \centering
     \begin{subfigure}[b]{0.49\textwidth}
         \centering         \includegraphics{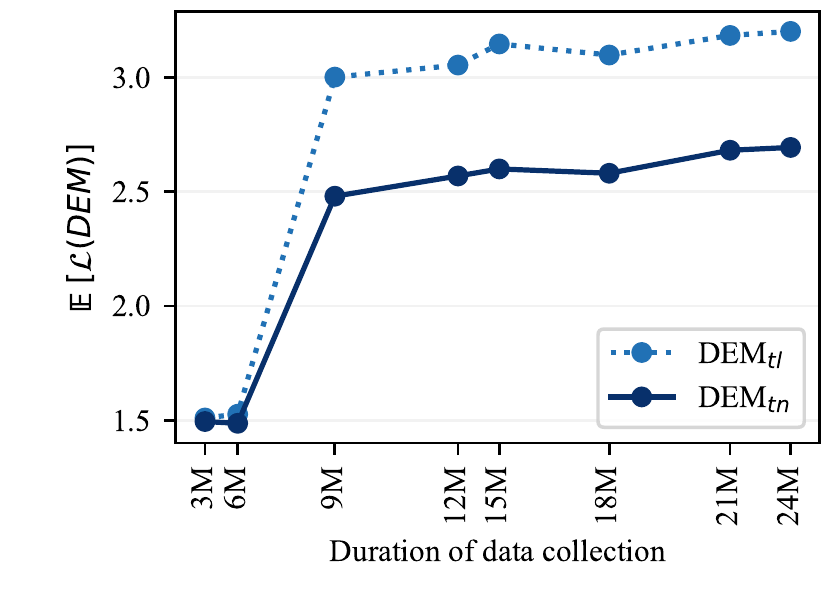}
         \caption{Expected log-likelihood of the DEM, $\mathbb{E}[\mathcal{L}(DEM)]$, represented as a function of the data collection period.}
         \label{fig:LL_vs_Trainsize}
     \end{subfigure}
     \hfill
     \begin{subfigure}[b]{0.49\textwidth}
        \centering  \includegraphics{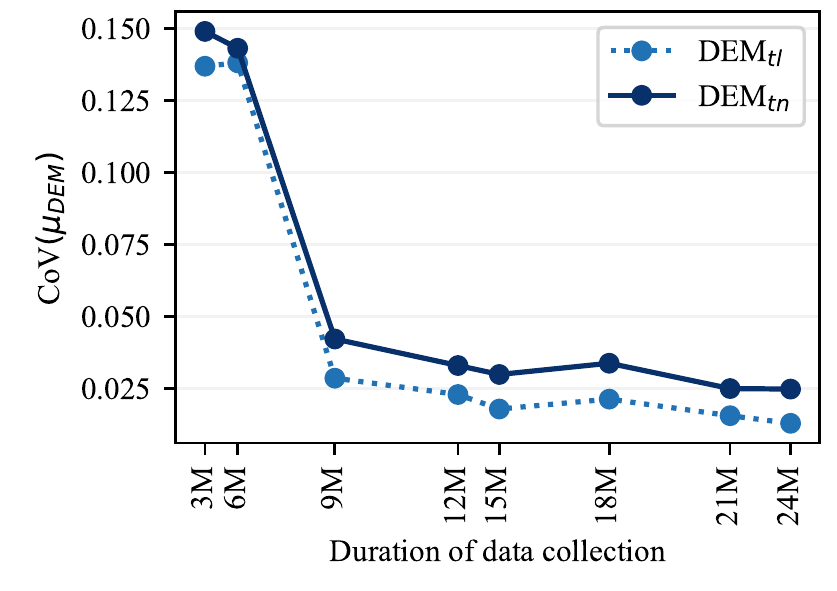}
        \caption{Normalized model uncertainty CoV$(\mu_{DEM})$ represented as a function of the data collection period.}
        \label{fig:CoV_vs_Trainsize}
     \end{subfigure}
     \caption{Illustration of the virtual monitoring model performance over specific data collection periods. It can be noticed that the amount of data collected differs for each period, e.g., the information retrieved over the course of the second trimester is scarce. Note that the plots indicate the mean of the model’s performance over the test dataset.}
     \label{fig:CoVLL_vs_Trainsize}
\end{figure*}

Following the process undertaken for the selection of a reduced set of input monitoring signals, we further investigate the uncertainty associated with the virtual load monitoring model with respect to specific data collection periods. For each conducted assessment, BNNs featuring the same architecture are trained and fed with data gathered throughout a certain period, i.e., from 3 to 24 months. Note that a test set randomly sampled from the longest available period, i.e., 25\% percentage of `fleet-leader’ full dataset, is used as reference to fairly evaluate all tested models.
More specifically, Figure \ref{fig:CoVLL_vs_Trainsize} represents the mean of the model’s performance over the test dataset and as a function of 3-month data collection periods. By examining the separation between tick values along the x-axis, one can notice the scarcity of monitoring data during certain periods, which might be potentially associated with the inactivity of certain sensors. 

In order to quantify the performance of the analyzed BNN models, two metrics are reported: the expected log-likelihood of the DEM, $\mathbb{E}[\mathcal{L}(DEM)]$, and the BNN's model uncertainty. The representation of the latter can be seen in Figure \ref{fig:CoV_vs_Trainsize}, which is computed over 10,000 forward model runs. For better interpretability, the model uncertainty, usually indicated by the standard deviation of the predicted means, SD$(\mu_{DEM})$, is normalized with respect to the expectation, $\mathbb{E}[\mu_{DEM}]$, as:
\begin{equation}
    \text{CoV}(\mu_{DEM}) = \frac{\text{SD}(\mu_{DEM})}{\mathbb{E}[\mu_{DEM}]},
    \label{eq:normalized_MU}
\end{equation}
where CoV$(\mu_{DEM})$ corresponds to (normalized) model uncertainty, and $\mathbb{E}[\mu_{DEM}]$ and SD$(\mu_{DEM})$ are computed according to Equations (\ref{eq:Expectation_mu}) and (\ref{eq:model_uncertainty}), respectively. Note that the standard deviation, SD(.), is equal to the square root of the variance, $\mathbb{V}(.)$. Even if the model uncertainty, i.e., CoV$(\mu_{DEM})$, is quantified from BNN predicted results, it does not directly assess the accuracy of the generated load predictions with respect to $DEM_{tl}$ and $DEM_{tn}$ (labels) measurements. The considered BNN models are, therefore, additionally evaluated with respect to the expected log-likelihood of load measurements, $\mathbb{E}[\mathcal{L}(DEM)]$, computed according to Equation (\ref{eq:expected_loglikelihood}) over also 10,000 forward model runs, and the results are plotted in Figure \ref{fig:LL_vs_Trainsize}. A similar trend in the performance reported by both metrics can be observed in the figure. 

In general, BNNs (and other deep learning models) will benefit from additional training data, especially if the information is collected for regions where previously available training data was limited. The amount of data required to establish a robust BNN is case-dependent as it is influenced by the number of neurons considered and the complexity of the inherent physical process, among others.
For the specific case of offshore wind turbines, representative environmental and operational data can be collected within a short term ($\sim$ 1-2 years). \cite{Santos2022,  Hlaing2022} As shown in Figure \ref{fig:CoVLL_vs_Trainsize}, the reduction of model uncertainty reported by the BNN steadily decreases over 12 months, where the model uncertainty reaches a stagnation point. One can thus conclude that, for this application, a robust BNN can be trained from 1-2 years of training data. Even if enough representative data has been collected, wind turbine dynamics might change at some point in the operational life, in which case, the BNN will automatically indicate to the user (e.g., operator) that additional data might be required. More specifically, an increased BNN's model uncertainty reported by the fleet-leader might suggest that the initially trained model is no longer adequate and strain gauges should be re-installed and the BNN model should re-trained.
Based on the reported findings, the BNN model is trained over the full training dataset, i.e., data collection spanning over 24 months, and the resulting BNN virtual model is then deployed to other wind turbines in the subsequent farm-wide monitoring study.

\subsection{Farm-wide deployment of virtual load monitoring model}

Once the BNN load model has been trained based on the monitoring output signals (labels) collected from the fleet-leader wind turbine, the resulting BNN model is deployed in order to predict the loads of the other two wind turbines in the same wind farm, denoted in the text as `MP01' and `MP02'. %Moreover, the load predictions estimated by the BNN model are compared against its deterministic counterpart. 
Following the same monitoring setup as the fleet-leader, the input standard monitoring signals fed to the network during the deployment stage correspond to the combination of SCADA and acceleration data collected at the level LAT-077 (top accelerometer), and the virtual load monitoring model predicts the damage equivalent moments at the level LAT-017. Throughout the study, the damage equivalent moments predicted for every 10-minute collected input data are randomly sampled, thus enabling an uncertainty quantification of the retrieved load predictions. 
Note that even if the predictions generated by the model are studied and deployed to only two wind turbines in this investigation, the proposed framework can be easily deployed to the whole wind farm. 

\begin{figure*}
    \centering    \includegraphics{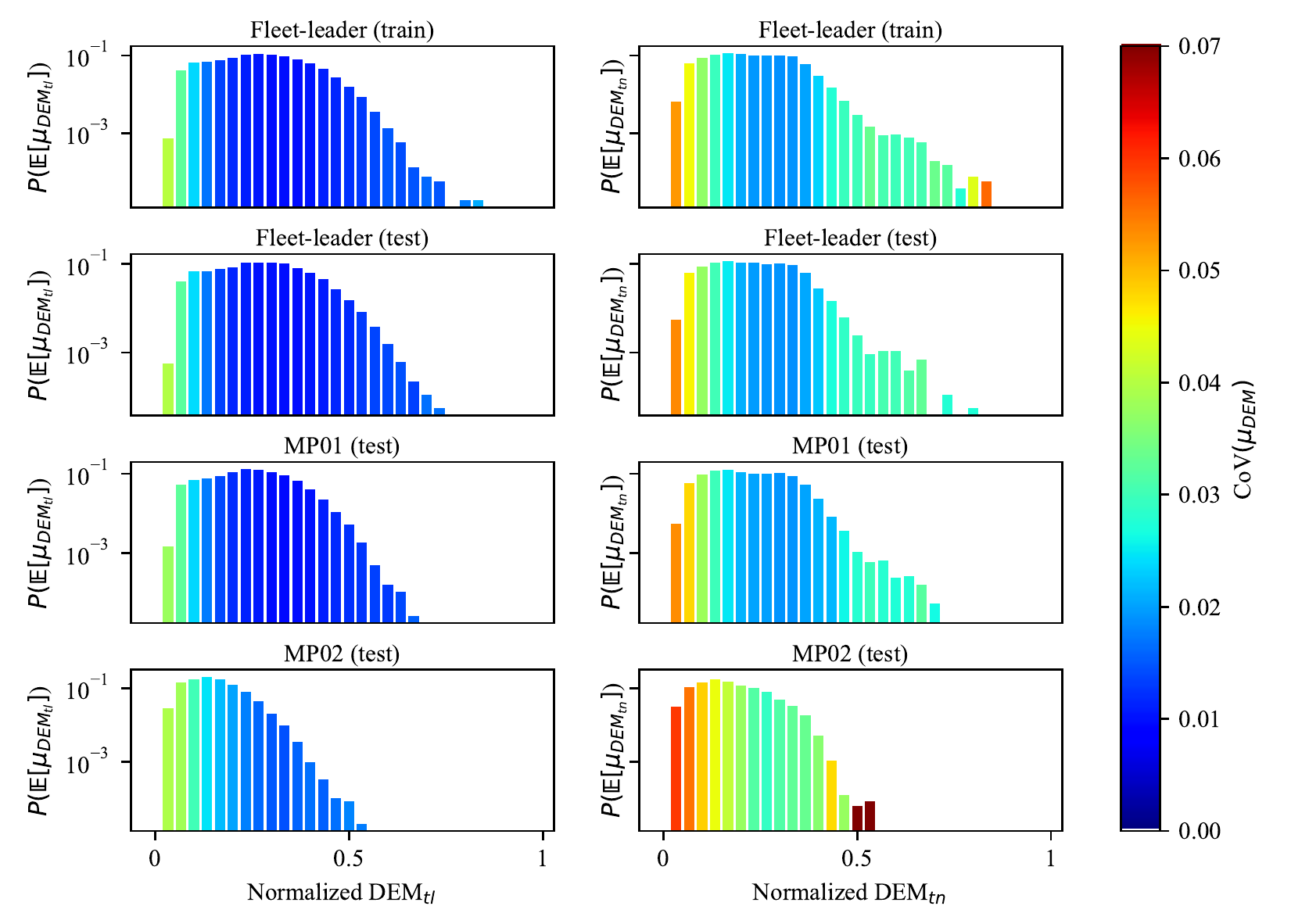}
    \caption{Load predictions generated by the Bayesian neural networks at the deployment stage for all analyzed offshore wind turbines. The retrieved expected damage equivalent moments, $\mathbb{E}[\mu_{DEM}]$, are classified into discrete bins colored according to their associated model uncertainty CoV$(\mu_{DEM})$. The height of each bar represents its probability and the color intensity indicates its associated model uncertainty.}
    \label{fig:DamageBelief_and_MU}
\end{figure*}

To observe the load predictions generated by the BNNs, Figure \ref{fig:DamageBelief_and_MU} illustrates the expected load indicator $\mathbb{E}[\mu_{DEM}]$ and its respective (normalized) model uncertainty CoV$(\mu_{DEM})$ for all investigated turbines. 
Specifically, the expected load indicators $\mathbb{E}[\mu_{DEM}]$ are allocated into DEM bins and normalized by the total number of samples, obtaining, therefore, a probability distribution over the damage equivalent moment. Note that $\mathbb{E}[\mu_{DEM}]$ is also equal to $\mathbb{E}[\hat{DEM}]$. The model uncertainty associated with each bin is then computed according to Equation (\ref{eq:normalized_MU}). As one can observe in the figure, the damage distribution is similar between the fleet-leader and MP01, resulting in higher probabilities in the low damage region (i.e.,  normalized $DEM < 0.5$) and smaller probabilities for medium to severe damage regions (i.e.,  normalized $DEM > 0.5$). However, MP02 rarely experiences medium to severe damage and many test samples are classified in the low damage region. Correspondingly, the model performance metric CoV$(\mu_{DEM})$ also indicates a good agreement between the fleet-leader’s training and testing, as well as MP01’s load predictions. The model in general predicts more accurately $DEM_{tl}$ loads than $DEM_{tn}$, achieving the lowest uncertainty in the medium $DEM_{tl}$ region of the fleet-leader and MP01 turbines. The model uncertainties reported for the case of MP02 are visibly higher than other turbines, announced in the illustration by a darker red color.

\begin{figure*}[h]
    \begin{subfigure}[b]{0.49\textwidth}
    \centering
    \includegraphics{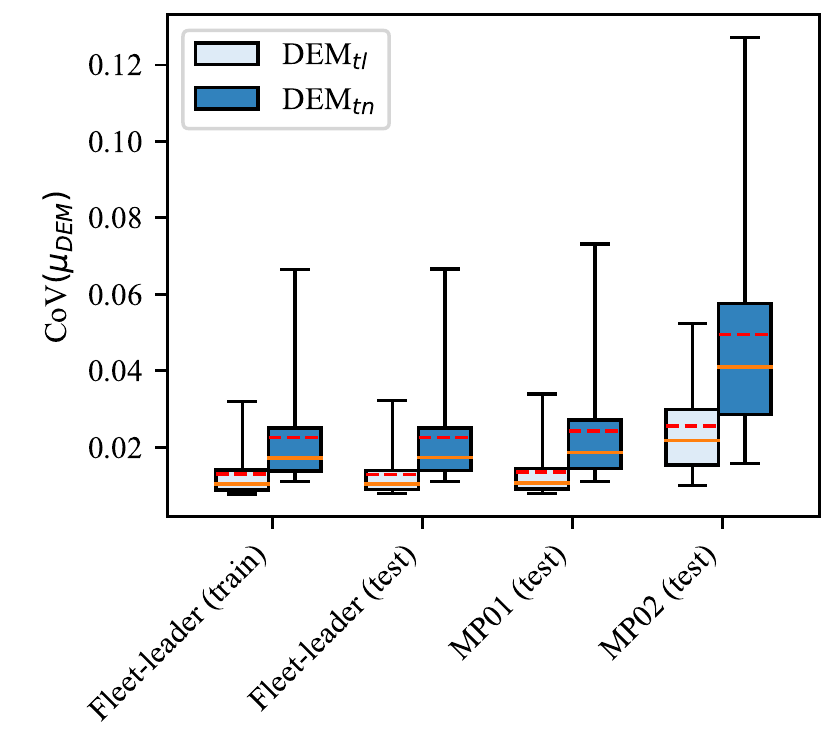}
    \caption{Model performance metric: Normalized model uncertainty announced by the implemented BNN, CoV$(\mu_{DEM})$.}
    \label{fig:MU_Farm}
    \end{subfigure}
    \begin{subfigure}[b]{0.49\textwidth}
    \centering    \includegraphics{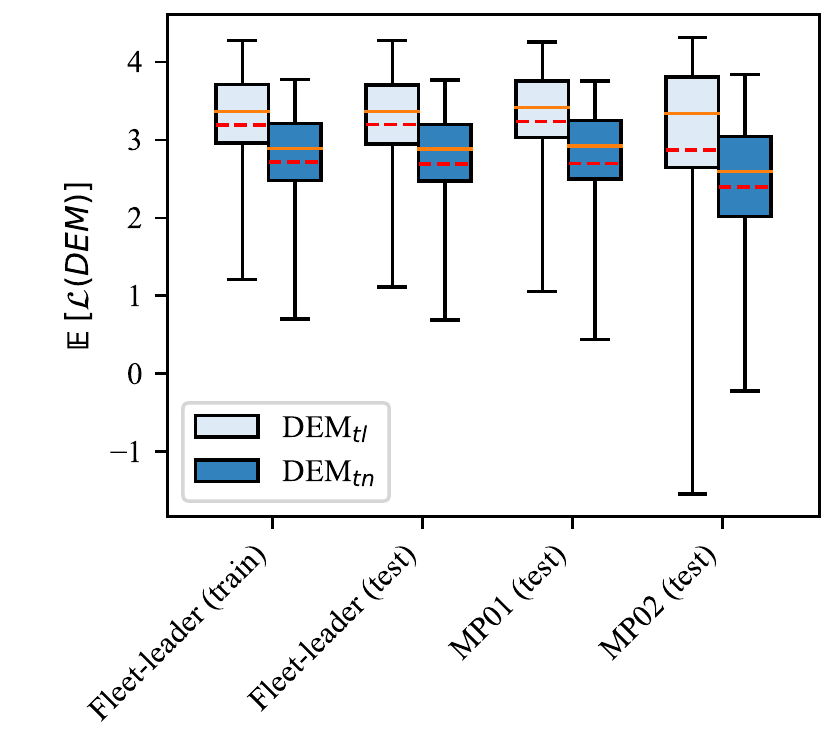}
    \caption{Model performance metric: Expected log-likelihood of the measured damage equivalent load, $\mathbb{E}[\mathcal{L}(DEM)]$.}
    \label{fig:LL_Farm}
    \end{subfigure}
    \caption{Representation of BNN’s model performance for farm-wide load prediction.  In particular, the model uncertainty is reported for the fleet-leader (both train and test datasets), MP01, and MP02 offshore wind turbines. %The metric used to indicate the model uncertainty $CoV(\mu_{DEM})$ corresponds to the standard deviation of the output mean statistical parameter, $SD(\mu_{DEM})$, normalized by the expected value of the output mean statistical parameter, $\mathbb{E}[\mu_{DEM}]$. 
    In the figure, the orange line and the red-dotted line represent, respectively, the median and mean values of CoV$(\mu_{DEM})$ over the corresponding dataset, and the boxes span between 25th and 75th percentiles, whereas the whiskers extend up to 2.5th and 97.5th percentiles.}
    \label{fig:performance_metric}
\end{figure*}

The showcased farm-wide load indicators and model uncertainty results are only based on the outputs retrieved from the deployed BNN, yet the accuracy of the predicted values with respect to the labels is not explicitly considered. Since MP01 and MP02 turbines have also been equipped with strain gauges during the monitoring campaign, the measurements (labels) are available to further analyze the obtained results. Figure \ref{fig:performance_metric} summarizes the model uncertainty for all considered offshore wind turbines, compared against the expected log-likelihood of the measured load indicator given the BNN outputs. Both performance metrics are represented by box plots that span over the interquartile range (IQR), i.e., between 25th and 75th percentiles, along with whiskers that extend up to 2.5th and 97.5th percentiles, respectively. Unsurprisingly, the expected log-likelihood outcomes are in agreement with the BNN-provided model uncertainties. MP02 wind turbine's results are characterized with higher model uncertainty as well as lower expected log-likelihood than for the other turbines, thus indicating potential conflicts with the generated predictions, e.g., the input monitoring data used during the training stage of the BNN might significantly differ from the input monitoring data collected for wind turbine MP02. 

\begin{figure}[h]
    \centering   \includegraphics{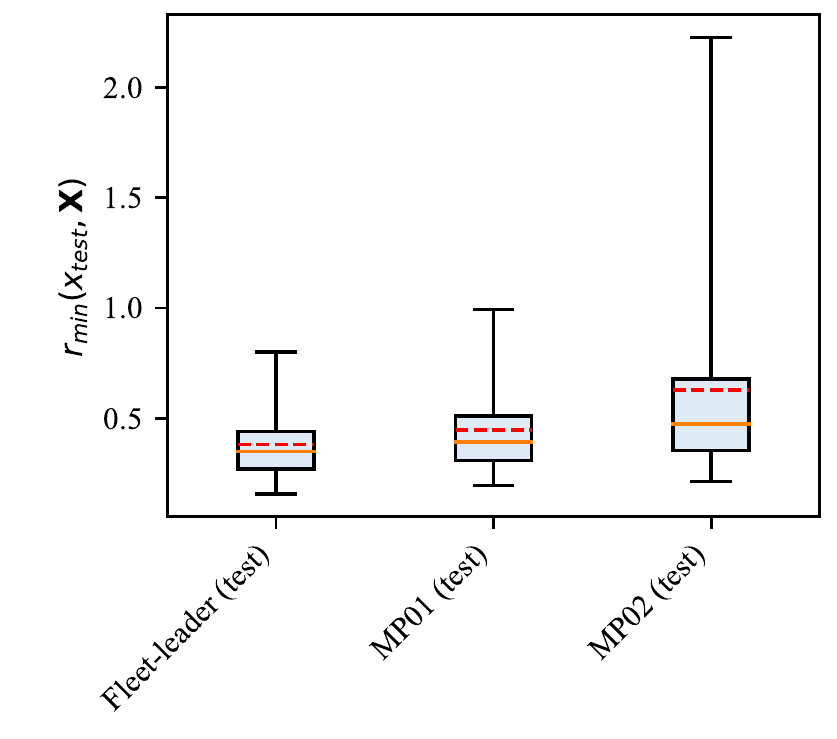}
    \caption{Representation of the minimum Euclidean distance from each wind turbine's input test dataset to the fleet-leader's input training dataset. The minimum Euclidean distances are plotted for the fleet-leader (test dataset), MP01, and MP02 offshore wind turbines. In the figure, the orange line and the red-dotted line represent, respectively, the median and mean values of $r_{min}({x}_{test},\boldsymbol{X})$ distances over their corresponding dataset, and the boxes span between 25th and 75th percentiles, whereas the whiskers extend up to 2.5th and 97.5th percentiles.}
    \label{fig:EU_distance_boxplot}
\end{figure}

To further clarify the potential differences between the input monitoring data used during the training of the fleet-leader and the input monitoring data available for the other wind turbines at the deployment stage, the minimum Euclidean distance, $r_{min}$, of the corresponding input test data, $\boldsymbol{x}_{test}$, with respect to the training dataset, $\mathbf{X}$, is also quantified:
\begin{equation}
   r_{min}(\boldsymbol{x}_{test},\mathbf{X}) =\min_{\boldsymbol{x} \in \mathbf{X}}|| (\boldsymbol{x}_{test}-\boldsymbol{x})||_2,
\end{equation}
where $\boldsymbol{x}, \boldsymbol{x}_{test}\in \mathbb{R}^{M}, \mathbf{X}\in \mathbb{R}^{M\times N}$, $M$ and $N$ stand for the dimension of input variables and the total number of training samples. $\mathbf{X}$ represents the matrix of input variables for all training data, $\boldsymbol{x}$ and $\boldsymbol{x}_{test}$ indicate the vectors of input variables for each training and testing sample, and $r_{min}$ denotes the minimum Euclidean distance from each test sample to its nearest training sample. A high $r_{min}$ value implies that the corresponding test sample is far from the training region and, consequently, the predicted results rendered by the model might be inaccurate or highly uncertain. As one can observe in Figure \ref{fig:EU_distance_boxplot}, MP01 turbine’s input monitoring test dataset is in good agreement with the fleet-leader’s train dataset, also characterized with similar Euclidean distances when compared with the fleet-leader test dataset. On the other hand, the observed high Euclidean distances with a wider spread over the test set demonstrate that MP02’s input test dataset substantially differs from the training dataset. Intrinsically, a BNN-based virtual monitoring scheme adjusts and reports higher model uncertainty, thus detecting potential conflicts that might emerge when input monitoring data at the deployment stage corresponds to unexplored data during the training stage. 

In particular, the lack of correspondence between training and MP02 testing datasets can be explained by their divergent structural dynamics behavior. To better visualize this, the marginal probability distributions corresponding to each input variable are compared for all tested turbines, and the results are presented in Appendix \hyperlink{AppendixB}{B}. Whereas SCADA input signals are fairly consistent, acceleration data clearly differ in MP02 wind turbine, as shown in Figure \ref{fig:Acc_disstributions}.
%However, it is worth-noting that the input space is high-dimensional, and the closer points in one dimension might have larger distance in the other dimensions. The actual distance between two points cannot be visualized by comparing each of the input distributions. 

\subsection{Comparative study between DNNs and epistemic BNNs}

\begin{figure*}[h]
     \centering
     \includegraphics{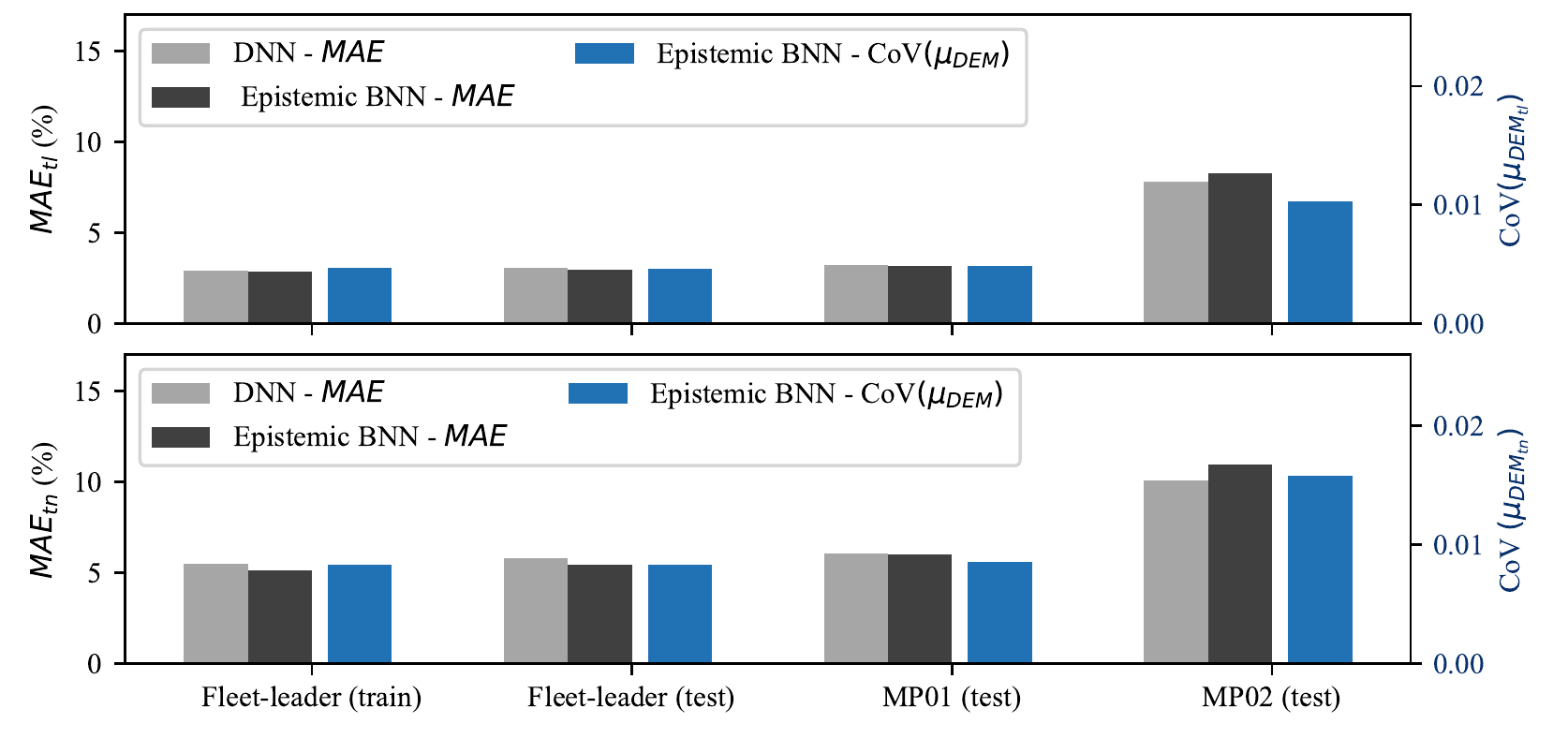}
     \caption{Prediction error associated with DNN and epistemic BNN predictions. The mean absolute error (MAE) corresponding to DNNs and epistemic BNNs is represented with light and dark grey bars, respectively. Additionally, model uncertainty metrics, $\text{CoV}(\mu_{DEM})$, reported by the epistemic BNN (without the need for ground truth labels) are represented with blue bars.}
     \label{fig:Percent_Model_error}
\end{figure*}

We have hitherto focused on the BNN model in which both aleatory and model uncertainties are encapsulated. This fully probabilistic BNN model is not directly comparable to the DNN since the former yields a probabilistic load estimate while the latter only generates a point estimate.
In order to objectively compare BNNs and DNNs in terms of error point estimates, an epistemic BNN (i.e., the variance of the prediction output is intentionally set up to 0) is trained based on the fleet-leader dataset and deployed to other wind turbines. In particular, this BNN modality quantifies the model uncertainty of the predicted load mean, yet the aleatory uncertainty is disregarded, as previously explained in the theoretical section. The outcome of the comparative analysis is represented in Figure \ref{fig:Percent_Model_error}, comparing the errors associated with DNN's deterministic outputs, $\hat{DEM}$, and BNN's predictions, $\mathbb{E}[\hat{DEM}]$, quantified following Equation (\ref{eq:percentage_error}). Note that the networks are trained only with the fleet-leader data and directly deployed to MP01 and MP02 without fine-tuning. The figure reveals that the epistemic BNN yields slightly more accurate point load estimates than its DNNs counterparts, except for MP02 turbine. Interestingly, the epistemic BNN also reports higher model uncertainty for MP02, illustrated in Figure \ref{fig:Percent_Model_error} with blue-colored bars, agreeing with the computed test accuracy (MAE). 

One should also keep in mind that the modeled epistemic BNN features fewer neurons than the considered DNN, potentially reducing its generalization capabilities. A thorough description of the implemented neural network architectures and training parameters is presented in Appendix \hyperlink{AppendixA}{A}. Whereas the generalization capabilities of different neural network architectures and hyperparameters are case-dependent, the proposed BNN-based virtual monitoring method constitutes a general framework for detecting potential load inaccuracies without the explicit need for a target (load measurement), which is especially relevant when dealing with farm-wide monitoring applications. In this specific case study, measurements are available for all analyzed turbines, yet this will most likely not be the case in practical scenarios as it is economically unfeasible to fully instrument all turbines. While the DNNs do not explicitly report model uncertainty estimates, BNNs are able to yield consistent model uncertainty information without the need for ground truth labels. This is confirmed by the reported results, where potential high prediction inaccuracy is automatically announced for the wind turbine MP02.

\subsection{BNNs model uncertainty}
BNNs are able to automatically announce the model uncertainty associated with the generated predictions independently of whether the output predictions are modeled as a probability distribution or as a point estimate. As mentioned in the theoretical section, a BNN that can capture both epistemic and aleatory uncertainty information would naturally require more training data. To analyze this, a further investigation has been conducted, comparing the model uncertainty reported by (i) a BNN that captures both aleatory and epistemic uncertainty components and (ii) an epistemic BNN (i.e., the variance of the prediction output is intentionally set up to 0). The architecture and training hyperparameters are the same for both tested BNNs. Further details can be found in Appendix \hyperlink{AppendixA}{A}.

\begin{figure}[htbp]
     \centering
        \includegraphics{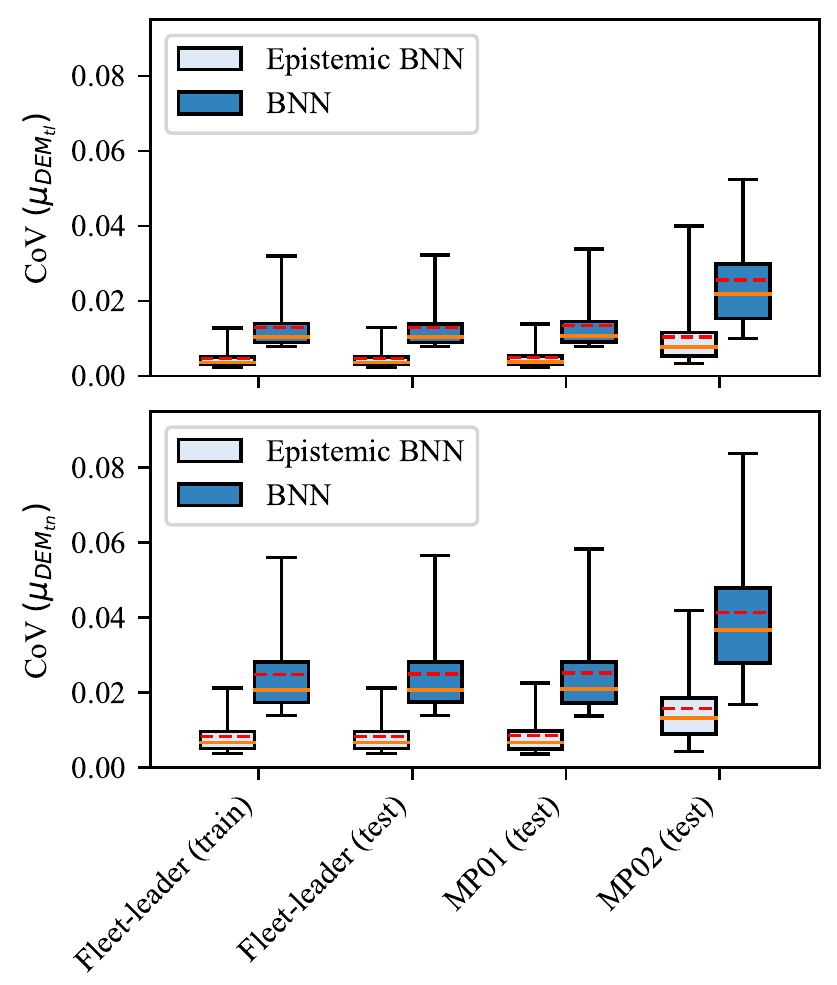}
         \caption{Model uncertainty associated with the load predictions generated by the investigated Bayesian neural networks. BNNs capturing only epistemic uncertainties are colored in light blue. Spreading over each wind turbine dataset, box plots represent the corresponding model uncertainty, $\text{CoV}(\mu_{DEM})$, within the interquartile range, with whiskers that span from 2.5th to 97.5th percentiles. Additionally, the median and mean are indicated with orange and red-dotted lines, respectively.}
         \label{fig:MUBoxplot_2BNNs}
\end{figure}

The results are represented in Figure \ref{fig:MUBoxplot_2BNNs}, indicating with box plots the spread in model uncertainty, i.e., $\text{CoV}(\mu_{DEM})$, over the training and testing datasets corresponding to all tested wind turbines. A similar trend can be observed for the analyzed BNNs, reporting a higher model uncertainty associated with the predictions generated for wind turbine MP02, thus verifying farm-wide applicability of both variants. A further inspection of Figure \ref{fig:MUBoxplot_2BNNs} reveals that the aleatoric BNN yields higher model uncertainty compared to its epistemic counterpart. This is justified by the fact that the training of a model that can predict a probability distribution given a certain input combination requires more training data, as mentioned previously, and especially considering the input is high-dimensional in this case, i.e., 13 input variables. Note that the prediction accuracy cannot be directly compared here because one of the tested BNN models generates an output probability distribution, whereas the epistemic BNN only provides the prediction of a point estimate.

\section{Conclusions}

In this paper, we propose and examine the effectiveness of a data-based virtual load monitoring framework formulated with Bayesian neural networks for `fleet-leader' farm-wide load monitoring, i.e., a data-based model is trained with data collected from a fully monitored wind turbine, and once the training task is completed, the resulting model is deployed, thus yielding load predictions for other non-fully monitored wind turbines. Within the investigation, we carefully assess relevant advantageous properties offered by Bayesian neural networks (BNNs), e.g., uncertainty quantification, automatic overfitting regulation, with respect to standard deterministic neural networks (DNNs), and we test the proposed virtual load monitoring framework through an experimental monitoring campaign conducted in an existing offshore wind farm. 

The results observed throughout the conducted experimental campaign reveal that, by reporting an epistemic model uncertainty indicator, and as opposed to their standard DNNs counterparts, BNN-based virtual monitoring approaches are able to intrinsically identify potential conflicts with the generated load predictions, providing therefore an informative proxy for controlling the accuracy of the deployed farm-wide virtual monitoring model. For instance, a BNN-based virtual monitoring model, whether aleatory uncertainty is inclusively modeled or not, will automatically report high model uncertainty during its deployment if the input monitoring data features outliers, i.e., unexplored data with respect to the data fed to the model during the training stage. Besides their useful uncertainty management capabilities, BNNs overall training task, even if more computationally demanding than for DNNs, is automatically regulated by Bayesian inference principles, thus avoiding the risk of overfitting and eluding the need of a separate cross-validation dataset.

From all available standard monitoring signals, a reduced set of informative input monitoring signals has been selected in this work by quantifying the generalization error resulting from each considered monitoring setup. The selection process relies, in this case, on already collected input data from the installed load monitoring system. Benefiting from BNNs’ internal properties in terms of uncertainty quantification, we motivate further research efforts toward farm-wide sensor placement studies capable of allocating monitoring installation and maintenance actions by following optimal adaptive management policies, e.g., asset management strategies identified via Markov decision processes and/or deep reinforcement learning methods. \cite{Morato2022, papakonstantinou2014planning, Andriotis2019, Morato2022a} 

In a more specific note, we also encourage the exploration of sophisticated principled metrics that can be computed during the testing stage of BNN-based virtual monitoring models, and we suggest, for instance, the investigation of entropy-based metrics. \cite{Depeweg2018a} Moreover, a detailed survey of data-based probabilistic virtual monitoring models is also recommended, e.g., comparing kernel- and neural network -based methods, which could potentially be complemented with a thorough uncertainty decomposition assessment.

\section*{Data availability}
Due to its proprietary nature and confidentiality concerns, supporting data of this research cannot be made publicly available. 

\section*{Acknowledgements}
This research is funded by the Belgian Energy Transition Fund (FPS Economy) through PhairywinD (\url{https://www.phairywind.be}) and MaxWind projects. We further acknowledge OWI-lab (\url{https://www.owi-lab.be}) for supporting with the wind farm dataset to develop this paper.

%\begin{thebibliography}{99}
%\bibliographystyle{SageH}
\bibliographystyle{SageV}
\bibliography{Bibliography.bib}
% %\end{thebibliography}

% \bibitem[Kopka and Daly(2003)]{R1}
% Kopka~H and Daly~PW (2003) \textit{A Guide to \LaTeX}, 4th~edn.
% Addison-Wesley.

% \bibitem[Lamport(1994)]{R2}
% Lamport~L (1994) \textit{\LaTeX: a Document Preparation System},
% 2nd~edn. Addison-Wesley.

% \bibitem[Mittelbach and Goossens(2004)]{R3}
% Mittelbach~F and Goossens~M (2004) \textit{The \LaTeX\ Companion},
% 2nd~edn. Addison-Wesley.

% \end{thebibliography}
% \newpage
~\newpage
\appendix
\hypertarget{AppendixA}{\section{Appendix A: Network topologies and training environments}}
\begin{table}[h]
\renewcommand\thetable{A1}
\small\sf\centering
\caption{Comparison between deterministic and Bayesian neural networks for sensor configurations  ``SCADA + wave, SCADA + wave + accelerometer (LAT-017), SCADA + wave + accelerometer (LAT-038), SCADA + wave + accelerometer (LAT-077), SCADA + wave + accelerometers (LAT-017, 038)".}
\begin{tabular}{lll}
\toprule
& Deterministic NN & Bayesian NN\\
\midrule
Optimizer & Adamax & Adam \\
& (lr=0.001)& (lr=0.0003) \\
Batch size & 32 & 1024 \\
No of episodes & 200 & 2000 \\
Early stopping & Validation loss & Training loss \\
 & (Patience = 5) & (Patience = 30)\\
No of neurons:\\
Hidden layers& 64, 128, 64& 31, 64, 32\\
Output layer& 2& 4\\
Distribution layer& -& 2\\
\bottomrule
\end{tabular}\\
\end{table}

\begin{table}[h]
\renewcommand\thetable{A2}
\small\sf\centering
\caption{Comparison between deterministic and Bayesian neural networks for sensor configuration  ``SCADA + wave + accelerometers (LAT-017, 038, 077)".}
\begin{tabular}{lll}
\toprule
& Deterministic NN & Bayesian NN\\
\midrule
Optimizer & Adamax & Adam \\
& (lr=0.001)& (lr=0.00035) \\
Batch size & 32 & 1024 \\
No of episodes & 200 & 2000 \\
Early stopping & Validation loss & Training loss \\
 & (Patience = 5) & (Patience = 30)\\
No of neurons:\\
Hidden layers& 64, 128, 64& 31, 64, 32\\
Output layer& 2& 4\\
Distribution layer& -& 2\\
\bottomrule
\end{tabular}\\
\end{table}

\newpage

\begin{table}[h]
\renewcommand\thetable{A3}
\small\sf\centering
\caption{Comparison between deterministic and Bayesian neural networks for sensor configurations  ``SCADA, SCADA + accelerometer (LAT-017), SCADA + accelerometer (LAT-038), SCADA + accelerometer (LAT-077)".}
\begin{tabular}{lll}
\toprule
& Deterministic NN & Bayesian NN\\
\midrule
Optimizer & Adamax & Adam \\
& (lr=0.001)& (lr=0.0002) \\
Batch size & 32 & 1024 \\
No of episodes & 200 & 2000 \\
Early stopping & Validation loss & Training loss \\
 & (Patience = 5) & (Patience = 30)\\
No of neurons:\\
Hidden layers& 64, 128, 64& 32, 64, 32\\
Output layer& 2& 4\\
Distribution layer& -& 2\\
\bottomrule
\end{tabular}\\
\end{table}

\begin{table}[h]
\renewcommand\thetable{A4}
\small\sf\centering
\caption{Comparison between deterministic and Bayesian neural networks for sensor configuration  ``SCADA + accelerometers (LAT-017, 038), SCADA + accelerometers (LAT-017, 038, 077)".}
\begin{tabular}{lll}
\toprule
& Deterministic NN & Bayesian NN\\
\midrule
Optimizer & Adamax & Adam \\
& (lr=0.001)& (lr=0.0003) \\
Batch size & 32 & 1024 \\
No of episodes & 200 & 2000 \\
Early stopping & Validation loss & Training loss \\
 & (Patience = 5) & (Patience = 30)\\
No of neurons:\\
Hidden layers& 64, 128, 64& 32, 64, 32\\
Output layer& 2& 4\\
Distribution layer& -& 2\\
\bottomrule
\end{tabular}\\
\end{table}

~\newpage
\onecolumn
\appendix
\hypertarget{AppendixB}{\section{Appendix B: Probability distributions of each input variable for the analyzed turbines}}
% \renewcommand\thetable{A\arabic{table}}  
% \setcounter{table}{0} 
% ~\newpage
% \appendix
% \hypertarget{AppendixB}{\section{Appendix B: Decomposition of the overall uncertainty for the analyzed wind turbines}}
% \renewcommand\thetable{A\arabic{table}}  
% \setcounter{table}{0} 

\begin{figure*}[h]
\renewcommand\thefigure{B1}
    \centering    \includegraphics{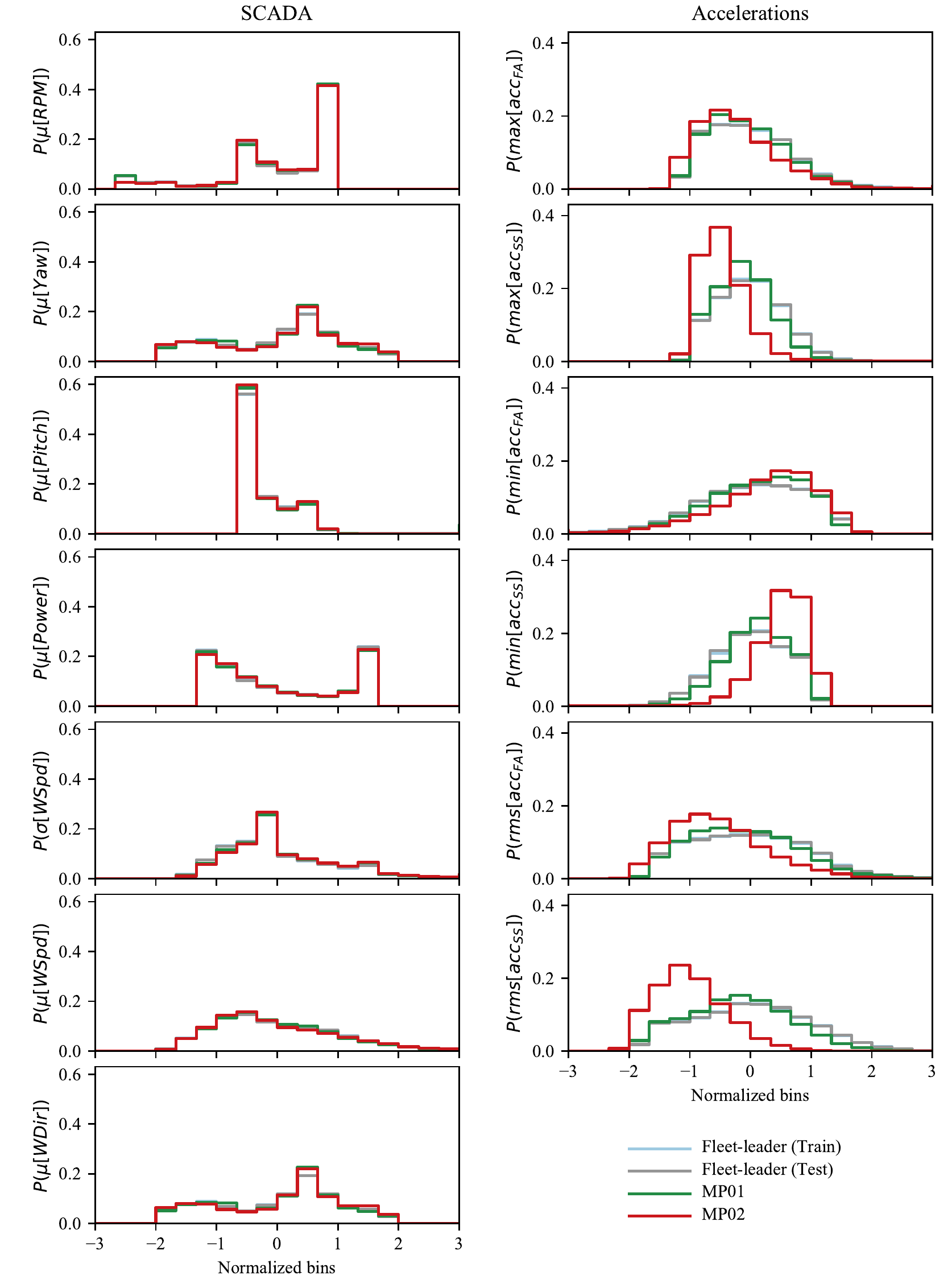}
    \caption{Probability distributions of 10-minute SCADA and acceleration statistics for the fleet-leader, MP01 and MP02 turbines. The X-axes are normalized due to data confidentiality concerns.}
    \label{fig:Acc_disstributions}
\end{figure*}

% \begin{figure*}[ht]
% \renewcommand\thefigure{B2}
%     \centering    \includegraphics{Figures/Scada_disstributions.pdf}
%     \caption{Probability distributions of 10-minute SCADA statistics the the fleet-leader, MP01 and MP02 turbines. The values of discretized bins are omitted due to confidentiality concerns.}
%     \label{fig:Scada_disstributions}
% \end{figure*}

\end{document}